# Multimodal remote sensing image registration with accuracy estimation at local and global scales

M.L. Uss[a], B. Vozel[b], V.V. Lukin[a], K. Chehdi[b]

[a]Department of Transmitters, Receivers and Signal Processing, National Aerospace University, Kharkov, Ukraine; [b]IETR UMR CNRS 6164 - University of Rennes 1, Enssat Lannion, France;

**ABSTRACT:**

This paper focuses on potential accuracy of remote sensing images registration. We investigate how this accuracy can be estimated without ground truth available and used to improve registration quality of mono- and multi-modal pair of images. At the local scale of image fragments, the Cramer–Rao lower bound (CRLB) on registration error is estimated for each local correspondence between coarsely registered pair of images. This CRLB is defined by local image texture and noise properties. Opposite to the standard approach, where registration accuracy is only evaluated at the output of the registration process, such valuable information is used by us as an additional input knowledge. It greatly helps detecting and discarding outliers and refining the estimation of geometrical transformation model parameters. Based on these ideas, a new area-based registration method called RAE (Registration with Accuracy Estimation) is proposed. In addition to its ability to automatically register very complex multimodal image pairs with high accuracy, the RAE method is able to provide registration accuracy at the global scale as covariance matrix of estimation error of geometrical transformation model parameters or as point-wise registration Standard Deviation (SD). This accuracy does not depend on any ground truth availability and characterizes each pair of registered images individually. Thus, the RAE method can identify image areas for which a predefined registration accuracy is guaranteed. This is essential for remote sensing applications imposing strict constraints on registration accuracy such as change detection, image fusion, and disaster management. The RAE method is proved successful with reaching subpixel accuracy while registering eight complex mono/multimodal and multitemporal image pairs including optical to optical, optical to radar, optical to Digital Elevation Model (DEM) images and DEM to radar cases. Other methods employed in comparisons fail to provide in a stable manner accurate results on the same test cases.

## 1. INTRODUCTION

Image registration (co-registration) is typical for many applications of remote sensing (RS) where one needs to bring two (reference and template) or more images of the same area taken by different means and in different conditions to the same coordinate system [1, 2]. Such a transformation allows fusion of multiview, multitemporal, multichannel, and multimodal images [3]. Images can also be registered (superimposed) to sensed terrain Digital Elevation Model (DEM) or digitized topographic map [4]. This provides pre-conditions for image joint analysis, extraction and dissemination of their informational content, terrain classification, and solving other typical RS tasks [5].

The main requirements to a good registration technique are the following. First, it should provide an appropriate, typically sub-pixel registration accuracy [6]. It is important to have some characterization of registration accuracy obtained at the output of the registration process for a given set of processed images (data) in order to be able to control the registration outcomes and to assure that they are appropriate for solving a considered task [7, 8]. Second, a method should be universal and reliable enough, i.e. it has to be applicable for different types of images and various kinds of underlying (sensed) terrains with minimizing probability of registration failure [9-11]. Finally, it is desirable to have a fully automatic registration which, at the same time, has to be simple enough to be realized in reasonable time [9, 11].

The main criterion for assessing a registration method performance is its registration accuracy. This accuracy can be described directly or indirectly by a variety of measures proposed in the literature [12]. The traditional criteria for assessing registration accuracy are MSE or RMSE as well as number and coverage of found correspondences. In practice, it is required to provide accuracy as high as possible for a given pair (or a set) of images. With this obvious and straightforward requirement, to quantify precisely accuracy of a registration method is, nevertheless, an open problem as it will be discussed in the next Section.

Concerning the second requirement, a universal registration method should be able to superimpose different types of data modalities typical for RS applications: optical-to-optical, optical-to-radar, radar-to-radar images as well as optical and radar to DEM or topographic data [4, 10, 11, 13, 14]. For each registration scenario and each registered pair of images, it is desirable either to provide registration accuracy required for a particular application or to specify that such an accuracy cannot be reached.

As for the third requirement, we concentrate below on fully automatic techniques. This does not mean that no *a priori* information is used. Many modern methods employ data on image corner geographic coordinates which are often available for exploited RS systems [15]. Besides, we pay attention to computation efficiency of considered and proposed techniques keeping in mind that complexity of a method can restrict its applicability.

Therefore, accurate, universal, and fully automatic registration with a controllable accuracy is the focus of our research. It is worth noting here that image registration methods have been under design for several decades [16] and, therefore, a lot of techniques have been already developed. Good surveys can be found in [1, 17]. Keeping this in mind, we feel that it is necessary to give a brief review of existing techniques that can be considered as the closest analogues of the method we propose in this paper (see Section 2). In turn, distinctive features of our method are the following. First, our method analyzes and controls potential registration accuracy at local and global scales with application to both linear and nonlinear geometrical transformations (Section 3). Due to this, one is provided with pixel-wise evaluated registration accuracy for a given pair of processed images without any ground truth available. Some aspects of processing acceleration are considered in Section 3 as well. Second, our method is fully automatic and universal. This is demonstrated in Section 4 where eight different cases of registration for various possible types of images (data) are studied and quantified. Comparisons to other techniques are also presented there. Third, our method is accurate and reliable. Its ability to provide high registration accuracy (subpixel or close to subpixel) is demonstrated alongside with ability to deal with complex registration cases where many other existing techniques fail to perform properly (see Section 4). Conclusions in Section 5 summarize the obtained results and discuss possible further directions of research.

## 2. BRIEF REVIEW OF EXISTING APPROACHES.

In this Section, we briefly review the problem of quantifying registration accuracy achieved by area-based or feature-based method. The difference between them is in direct use of images intensities in the registration process for area-based methods (also called intensity-based methods) or differently an "informational extract" from these intensities, called features and possessing some degree of invariance to image formation conditions (illumination, geometrical distortions, modality) for feature-based method. We also recall existing approaches for multimodal image registration and reported accuracy of such methods.

This paper considers the widespread and modern registration approach [1] for which registration at the global image scale is fully based on correspondences between pairs of registered image fragments detected at the local scale using either feature-based or area-based methods. Following this approach, to quantify registration accuracy at the global scale, one should be able to determine registration accuracy of each eligible correspondence.

Concerning feature-based methods, solutions exist for simple point-like features only [18]. For complex features like SIFT (Scale-Invariant Feature Transform), Harris corners or objects contours, no solutions have been published yet to the best of our knowledge.

Using area-based methods, local correspondences are mainly found by optimizing a suitable similarity measure [19]. Potential registration accuracy achievable using the sum of squared differences (SSD) measure was derived in [7] and [8] in the form of Cramér–Rao lower bound (CRLB) on estimation error of geometrical transform parameters. As it has been shown in [20], this solution does not reflect correctly the dependence of potential registration accuracy on signal-to-noise ratio (SNR) and correlation between reference and template images. Thus, it cannot be considered for application in multitemporal and multimodal cases. Characterization of registration accuracy further complicates for more sophisticated similarity measures like Normalized Correlation Coefficient (NCC), Mutual Information (MI) or Phase Correlation.

Without knowing registration accuracy at the local scale, it is impossible to obtain registration accuracy at the global scale, i.e. estimation accuracy of parameters of geometrical transformation model between reference and template images. For simple point-like features, solutions for registration accuracy at the global scale were derived in [8, 21]. The problem with these approaches is that they do not take into account outliers among correspondences, a situation rarely met in practice.

The MLESAC (Maximum Likelihood Estimation Sample Consensus) registration method initially developed by Torr and A. Zisserman in [22] and recently upgraded by J. Ma *et al.* and J. Zhao *et al.* in [23, 24] comes furthermost in the direction of an adequate use of features registration accuracy. It solves the registration problem by optimizing the complete likelihood function via EM (Expectation Maximization) approach, and by including registration accuracy of features (e.g., SIFT) in the score function. However, in this approach, feature registration accuracy does not come directly from the feature properties analysis. Instead, it follows as an *a posteriori* estimate at the registration process output. This estimate is the same for all features and, thus, it does not characterize specifically individual features that may exhibit significantly different properties. Let us note that the registration accuracy at the global scale was not derived in [22-24]. Similarly, registration error of point correspondences is characterized as variance of residual registration noise in [25]. This variance is used at the outlier rejection stage and takes the same value for all correspondences.

The methods mentioned above illustrate a common approach that consists in characterizing registration accuracy *a posteriori* at the output of registration process, mainly caused by inability to quantify registration accuracy at the local scale. For instance, if registration accuracy of an individual feature cannot be directly derived from the analysis of registered images properties, one has to analyze feature position deviation from a ground truth. For research purpose, the ground truth can be obtained from collected ground control points (e.g. GPS based). Here, we assume a more realistic situation with no ground truth available. In this situation, registration accuracy of the features is typically inferred by analyzing their deviation from the estimated geometrical model (e.g. utilizing leave-one-out approach). We use the term *a posteriori* to underline the dependence of the analysis carried out on the registration output. While such a reasonable analysis gives an idea of a registration method performance, it has two serious drawbacks. First, registration accuracy of a method evaluated for one pair of images does not characterize its performance for other pairs. Indeed, registration accuracy depends on similarity of the registered images that may vary significantly from one pair to another one (e.g., due to registration problem type: either monomodal, mutlitemporal or multimodal), or within registered images as a result of cloud cover influence or different properties of land cover types like urban, rural, forest, water surface, etc. Second, this *a posteriori* registration accuracy does not support registration process as it is unknown (unavailable) at the beginning of registration process.

The goal of this paper is to investigate more deeply the registration accuracy of remote sensing images both at the local and global scales so as to highlight advantages that can be drawn from using this additional information in the registration process. For this purpose, we associate potential registration accuracy with each correspondence between pairs of image fragments where this accuracy is intrinsically linked with properties of registered images fragments. Potential registration accuracy is then utilized in the registration process as additional *a priori* information and it also supports outliers/inliers detection stage. Finally, at the output, we quantify registration accuracy at the global scale with no longer use of ground truth or manual control points, possibly available.

The possibility to move forward in the direction of quantifying registration accuracy comes essentially from the Cramér–Rao Lower Bound $CRLB_{fBm}$ on image registration errors first introduced in [20] for pure translation model, and further developed in our paper [26] for rotation-scaling-translation (RST) model. Recently, the $CRLB_{fBm}$ bound has been extended to the case of optical-to-radar image registration by accounting for spatial correlation properties of speckle noise and its strong signal-dependency [27]. An interesting advantage of the $CRLB_{fBm}$ bound over state-of-the-art alternatives is that it can be applied to evaluate registration accuracy of correspondences found by the area-based approach for all kinds of registration problems including monomodal, multiview, multitemporal and multimodal cases. At the local scale, it takes into account reference and template image fragments texture and noise properties such as texture amplitude and roughness, correlation between reference and template images, noise spatial correlation and signal-dependency. This confers to the $CRLB_{fBm}$ bound the capability for accurately predicting registration accuracy of known area-based registration methods including NCC, MI, phase correlation [20, 26, 27].

The main contribution of this paper is a new area-based image registration scheme, called RAE (**R**egistration with **A**ccuracy **E**stimation) that beneficially involves registration accuracy at all stages, including the preliminary search of putative correspondences (PC), followed by outlier detection and estimation of geometrical transformation parameters. At the local scale, we assign the registration accuracy estimated using the $CRLB_{fBm}$ bound to each PC between control fragments (CF) of registered images. At this stage, main properties of texture and noise of a registered pair of image CFs are taken into account. Registration accuracy is directly exploited to range PCs in order of increasing contribution to registration. At the outlier detection stage, leave-one-out cross-validation (LOOCV) approach is employed: actual position of each PC is compared to its prediction based on other PCs. Detection is done by comparing the error of PC position prediction with a threshold. The threshold for each PC is calculated using PCs registration accuracy previously derived. At the stage of estimating geometrical transform parameters, covariance matrix of parameter estimates is obtained from the derived PCs registration accuracy and outlier detection results. At the output, we provide standard deviation (SD) of registration error for each pixel of the reference image. This registration error is individual for each pair of the registered image fragments. It reflects more adequately the structure of the registered images that is composition of areas suitable for registration in higher or less degree: urban, rural, forest, cloud cover, etc., and noise properties of both reference and template images.

For the sake of clarity, let us summarize elements of our previous works that are inherited by the newly proposed RAE method:

1. The $CRLB_{fBm}$ lower bound on RST parameters estimation error derived in [20] and [26];

2. Estimator of the fBm-field parameters from noisy fragments of the reference and template images aligned with RST model [26];

3. Extension of the $CRLB_{fBm}$ to the case of more practically realistic noise model including signal-dependency and non-negligible spatial correlation [27]. This noise model is essential for multimodal registration cases, especially when radar and DEM images are involved;

The core of the proposed RAE method is a novel criterion for joint outlier detection and estimation of registration parameters. We extend the criterion proposed by Ma *et al* in [23] to better take into account the estimated registration accuracy of each PC and allow a single template CF to have multiple PCs to the reference image fragments. Other two improvements allow us, on one hand, to get rid of additional constraint on geometrical transformation, namely, local linearity, and, on the other hand, to reformulate outlier detection from the point of view of LOOCV approach. The new criterion we propose assures that the found correspondences do not contain high-leverage points [28]. We will demonstrate that all these improvements allow performing successful registration for general affine and second-order polynomial models when probability of finding the true correspondences is as low as 2%.

Asymptotically, the RAE method is characterized by linearithmic complexity with respect to image size (number of control fragments). But, in practice, it has high computational complexity as it relies on $CRLB_{fBm}$ bound. We reduce this complexity by an incremental scheme where each newly found correspondence between reference and template image fragments shrinks the search zone in the geometrical transform parameter space. With this approach, we are able to reach acceptable registration time.

Let us underline that the input of RAE method is a set of PCs found by any area- or feature-based approach provided the registration accuracy of a method used can be precisely quantified, i.e. there is a way to predict registration error standard deviation directly from image properties (signal-to-noise ratio, noise signal-dependency, noise spatial correlation, texture roughness, structural properties etc.) without performing actually a registration with this method. In this paper, we follow the area-based approach using the NCC similarity measure [19, 29, 30] and intensity interpolation for reaching subpixel registration accuracy. The choice of the NCC method is justified by the fact that its performance is rather close to the $CRLB_{fBm}$ bound in those image areas where $CRLB_{fBm}$ is the most suitable, i.e. isotropic textures with normal increments [26]. In the experimental part of the paper, the capabilities of the proposed RAE method using NCC will be demonstrated by solving the most complex registration problems including radar-to-optical, optical-to-DEM and DEM-to-radar real multimodal scenarios in a unified manner.

The first case, optical-to-radar registration, is a well studied problem for which two dominant approaches exist at the moment: either methods utilizing MI similarity measure or feature-based methods [31]. Registration of high resolution optical to radar images in urban areas using MI similarity measure with additional segmentation step was studied in [10]. The method can only deal with pure translation estimation. The reported registration RMSE is from 0.96 to 2.6 pixels for large control fragments of size 300 by 300 pixels. MI measure was used in [11] to find and localize correspondences between optical and radar images at the local scale. The drawback of the method is that this method finds a very limited number of correspondences.

L. Hui *et al.* explored contour-based approach to optical-to-radar image registration in [13]. The method applies chain-based correlation of closed contours and salient features of open contours to estimate parameters of RST transform between optical and radar images. The reported RMSE of control points is about 1.1…2.1 pixels. However, the method can be applied only when initial registration error is small (less than about 5 pixels). A descriptor called shape-context is proposed for optical-to-radar image registration in [32]. It is based on distribution of edge features in log-polar space. For this method, the reported RMSE of the registered control points is about 1.8 pixels. Classical SIFT descriptor has been found not suitable when directly applied to remote sensing images in general [33] and to optical and radar images in particular [11]. Improvements to SIFT descriptor were proposed by Suri *et al*. in [34] and further developed by Bin *et al.* in [31]. A mixed approach utilizing MI method at the coarse-registration stage and line features at the fine registration stage [14] demonstrated registration RMSE of 5 pixels. Overall, the reported RMSE of control points for optical-to-radar image registration varies from 1 to 5 pixels [4, 10, 14, 31, 32]. It is interesting to note that the lower boundary of this interval – 1 pixel – was identified as a lower limit for optical-to-radar images registration accuracy (RMSE of correspondences) in [27]. In this paper, we will demonstrate that the RAE method is able to provide RMSE of about 0.75…1.05 pixels and subpixel registration accuracy at the global scale for images covering rather featureless areas without large-scale water objects where other methods (used in comparisons) either provide less accurate results or fail.

Optical-to-DEM registration is even more complex (we refer interested readers to experimental section of [35] for discussion of such a registration scenario complexity). Successful optical-to-DEM image registration based on contour-based approach and Non-Uniform Rational B-Splines (NURBS) was reported in [4]. The method can deal with both affine and perspective geometrical models and the achieved RMSE of control points is 2.3…2.8 pixels. The drawback of this method is in the usage of manual segmentation stage and inability to provide control points in image areas without rivers or other water object boundaries. Murphy and Le Moigne in [35] utilized shearlet-based features to register multimodal RS data under affine transformation. This approach was reported to register optical to shaded DEM images with mean RMSE of about 3.5 pixels but for optical-to-DEM case it was not successful. In turn, for the optical-to-DEM registration problem, our RAE method provides control points present at different land covers with RMSE of about 0.62…0.75 pixels. Similarly to optical-to-radar registration case, subpixel registration accuracy at the global scale is achieved.

The last, DEM-to-radar registration scenario is of even higher level of complexity and, to the best of our knowledge, no successful registration cases were reported in the literature. Our RAE method can handle this registration case with the RMSE of found control fragments of about 0.67…0.85 pixels.

# 3. IMAGE REGISTRATION METHOD UTILIZING POTENTIAL REGISTRATION ACCURACY AT LOCAL AND GLOBAL SCALES

This Section formally introduces the proposed RAE registration method and discusses its main features. We start by recalling the constraints that can be considered for modeling a geometrical transformation at both local and global scales for remote sensing applications. Then, we define and state the problem of interest. We describe the search for putative correspondences using NCC similarity measure, assignment of potential registration accuracy to each found PC, outlier detection, and estimation of registration parameters at the global scale. Then, registration accuracy at the global scale is derived. Lastly, computational complexity of the RAE method is analyzed in detail.

**3.1. Constraints on geometrical transformation model parameters for remote sensing applications**

Geometrical errors encountered in RS applications have their own specificities. Due to remote sensors' linearity (initially assured in their design), affine transformation hypothesis can be accepted as a first-order approximation of geometrical transformation between two RS images: the errors nonlinearity caused by sensor deviations from ideal linear array camera model, Earth curvature, etc. are limited as compared to the main linear part of the sensor behavior. A preponderant error source is translation error due to errors in satellite positioning and orientation. Scale and orientation errors are significantly smaller and can be locally treated as spatially varying translation errors (drift errors). Nonlinear distortions can be most of the time neglected at the local level [15]. As a result, at the local scale, transformation between reference and template images reduces itself to rotation-isomorphic scaling-translation (RST) model with smaller estimation errors w.r.t. rotation and scaling and outweighing translation estimation errors [33]. Such assumptions can be considered typical for RS images registration field [31, 36].

Initial geopositioning of RS images is generally based on a rigorous sensor orbital model. The registration error provided with this method (no ground control points available) is called direct geopositioning error and it defines initial search zone w.r.t. translation components. The value of direct geopositioning error reduces gradually with advances in technology. But this process is accompanied by improvement of sensors spatial resolution [5]. For example, for OLI sensor of Landsat 8 satellite, direct geopositioning error (CE90 or 2 sigma) is 65 m. Expressed in pixels (15m spatial resolution for PAN band), this gives the error of 6.5 pixels (3 sigma). Direct geopositioning error is 56.56 for QuickBird/QuickBird-2, 16.50 for IKONOS, 15 for Radarsat-2, and 3.81 pixels for RapidEye satellites (all values mentioned here were calculated using sensor specifications). Overall, direct geopositioning error expressed in pixels is more or less a stable value with range of variation from 3 to about 60 pixels (3 sigma) (from a few pixels to a few tens of pixels according to [35]).

In this paper, we assume that images to register are orthorectified (in the experimental part we use preliminary orthorectified images or perform relief correction using DEMs for study areas). Under these conditions, we assume a polynomial model as geometrical transformation model at the global scale. The proposed RAE method is applicable for arbitrary model order. In the experimental part of the paper, we provide results for first- and second-order models. More accurate geometrical modeling is, however, possible using Rational Polynomial Coefficient (RPC) model. We prefer here to consider a simpler polynomial model to better illustrate the proposed method advantages leaving RPC model for future study.

At the local scale, the RST model is chosen as justified above. Initial values of the model parameters are estimated from metadata provided with the reference and template images (box corners coordinates). The search zone w.r.t. translation parameters is set based on direct geopositioning error of the reference and template image platforms (the procedure is described in the next subsection). Designing the RAE method, we pursue the goal to assure robust registration for quite large initial translation errors up to ±100 pixels in order to cover direct geopositioning errors commonly met in practice.

**3.2. Definitions, constraints and problem statement**

Two images to register are the reference image (RI) with $m_{RI}$ rows and $n_{RI}$ columns and the template image (TI) with $m_{TI}$ rows and $n_{TI}$ columns. Coordinates of a pixel in $i$th row and $j$th column of the RI/TI images will be referred to as $\mathbf{y} = (i_{RI}, j_{RI})^T$ and $\mathbf{x} = (i_{TI}, j_{TI})^T$, respectively.

The reference and template images are initially coarsely registered based on longitude and latitude of all four corners embedded in the imagery files metadata. The initial registration is, thus, described by the affine transformation:

$$\mathbf{y}_{\text{Init}} = \mathbf{A}_{\text{Init}} \mathbf{x} + \mathbf{d}_{\text{Init}}. \tag{1}$$

An affine transformation matrix, $\mathbf{A}$, can be approximated by RST transformation matrix $\mathbf{A}_{\text{RST}} = \Delta r \mathbf{R}$, where $\Delta r$ is a scaling factor and $\mathbf{R} = \begin{pmatrix} \cos(\alpha) & \sin(\alpha) \\ -\sin(\alpha) & \cos(\alpha) \end{pmatrix}$ is a rotation matrix through an angle $\alpha$. Optimal values of $\Delta r$ and $\mathbf{R}$ are found according to the method described in [23] (see Section C "Rigid Feature Matching" and references therein): $\Delta r = \sqrt{|\mathbf{A}|}$ and $\mathbf{R} = \mathbf{U}\mathbf{V}^T$, where $\mathbf{A} = \mathbf{U}\mathbf{S}\mathbf{V}^T$ is SVD decomposition of $\mathbf{A}$ and $\mathbf{A}$ does not include reflections. The initial value for the rotation angle $\alpha_{\text{Init}}$ and the scaling factor $\Delta r_{\text{Init}}$ between RI and TI images are estimated by applying this decomposition to $\mathbf{A}_{\text{Init}}$.

Initial registration error with respect to both spatial coordinates is bounded above by $d_{\max 0}$: $\sqrt{(i_{\text{RI.Init}} - i_{\text{RI0}})^2 + (j_{\text{RI.Init}} - j_{\text{RI0}})^2} \leq d_{\max 0}$, where $\mathbf{y}_0 = (i_{\text{RI0}}, j_{\text{RI0}})$ are the coordinates of the true correspondence, $\mathbf{y}_{\text{Init}} = (i_{\text{RI.Init}}, j_{\text{RI.Init}})$ is given by (1). The value of $d_{\max 0}$ can be set based on direct geopositioning error standard deviations (SD), $\sigma_{\text{g.RI}}$ and $\sigma_{\text{g.TI}}$, of the reference and template images, respectively: $d_{\max 0} = 3\sqrt{\sigma_{\text{g.RI}}^2 + \sigma_{\text{g.TI}}^2 \Delta r_{\text{Init}}^2}$, where $d_{\max 0}$ and $\sigma_{\text{g.RI}}$ are given in RI pixels and $\sigma_{\text{g.TI}}$ in TI pixels. The values of $\sigma_{\text{g.RI}}$ and $\sigma_{\text{g.TI}}$ can be found in specifications of sensors which the RI and TI images were acquired with.

As it has been introduced above, we consider the geometrical transformation model at the global scale in the polynomial form of degree $n$:

$$\mathbf{y} = \mathbf{g}(\mathbf{x}, \mathbf{c}_1, \mathbf{c}_2), \tag{2}$$

where $\mathbf{g}(\mathbf{x}, \mathbf{c}_1, \mathbf{c}_2) = (g(\mathbf{x}, \mathbf{c}_1), g(\mathbf{x}, \mathbf{c}_2))$, $g(\mathbf{x}, \mathbf{c}) = g(i_{\text{TI}}, j_{\text{TI}}, \mathbf{c}) = \mathbf{c} \cdot (1, i_{\text{TI}}, j_{\text{TI}}, \ldots, i_{\text{TI}}^{k_1} j_{\text{TI}}^{k_2}, \ldots, i_{\text{TI}}^n, j_{\text{TI}}^n)$,

$0 \leq k_1 + k_2 \leq n$, $\mathbf{c}$ is $n_c = (n+2)(n+1)/2$ column vector of coefficients, $\mathbf{c}_1$ and $\mathbf{c}_2$ define a transformation with respect to horizontal and vertical directions. Initially, $\mathbf{c}_1 = (\mathbf{d}_{\text{Init}}(1), \mathbf{A}_{\text{Init}}(1,1), \mathbf{A}_{\text{Init}}(1,2))$ and $\mathbf{c}_2 = (\mathbf{d}_{\text{Init}}(2), \mathbf{A}_{\text{Init}}(2,1), \mathbf{A}_{\text{Init}}(2,2))$ for $n=1$; $\mathbf{c}_1 = (\mathbf{d}_{\text{Init}}(1), \mathbf{A}_{\text{Init}}(1,1), \mathbf{A}_{\text{Init}}(1,2), 0, ..., 0)$ and $\mathbf{c}_2 = (\mathbf{d}_{\text{Init}}(2), \mathbf{A}_{\text{Init}}(2,1), \mathbf{A}_{\text{Init}}(2,2), 0, ..., 0)$ for $n>1$.

For both registered images, relief influence is taken into account by introducing systematic correction factors, $\Delta i(i, j, H(i,j))$ and $\Delta j(i, j, H(i,j))$, compensating the observed shift at point $(i, j)$ due to relief with height $H(i,j)$ (obtained from DEM for the study area).

The $CRLB_{\text{fBm}}$ bound is currently restricted to the RST model; therefore, we should constraint transformation (2) to be well approximated by the RST model at the local scale (about ±10 pixels for the proposed method). This constraint is natural for RS sensors as it has been discussed in subsection 3.1. In order to impose it mathematically, model (2) is first approximated by an affine transformation $\mathbf{y} = \mathbf{A}_{\text{linear}} \mathbf{x} + \mathbf{d}_{\text{linear}}$ in the neighborhood of a point $\mathbf{x}_0 = (i_{\text{TI0}}, j_{\text{TI0}})$ using first-order terms of its Taylor series expansion. We checked that for all test cases considered in the experimental part of the paper, nonlinear effects at the distance of 10 pixels are of the magnitude order of $10^{-2}$ pixel, that can be reasonably neglected.

Then, the affine transformation matrix $\mathbf{A}_{\text{linear}}$ is further approximated by the RST transformation matrix $\mathbf{A}_{\text{RST}}$. In the neighborhood of $\mathbf{x}_0$, uncompensated difference between RI and TI fragments at the RI coordinate system can be measured through matrix difference $\mathbf{A}_{\text{linear}} - \mathbf{A}_{\text{RST}}$. Maximum uncompensated error at the one pixel distance, defined as $d_1$ is obtained as the square root of the maximum eigenvalue of matrix $(\mathbf{A}_{\text{linear}} - \mathbf{A}_{\text{RST}})(\mathbf{A}_{\text{linear}} - \mathbf{A}_{\text{RST}})^T$. For a RI fragment of ±10 pixels size, this error increases to $d_{10} = 10 d_1$. For image pairs considered in this paper, we found that $d_{10}$ vary from 0.03 to 0.3 pixels. This error can be neglected as it is tolerated by area-based methods, such as the NCC method in our case [37, 38].

With these definitions, the registration problem is formulated as an estimation problem of parameter vectors $\mathbf{c}_1$ and $\mathbf{c}_2$.

**3.3. Search for putative correspondences**

Let us tile the template image by non-overlapping CFs of size $N_{TI} \times N_{TI}$ pixels and define the center coordinates of these CFs by $\mathbf{x}_k$, $k = 1...N_{CF}$ where $N_{CF}$ is the total number of CFs.

Using the current estimate of transformation parameter vectors $\mathbf{c}_1$ and $\mathbf{c}_2$ (see Fig. 1), position of the center of the transformed template CF is predicted as $\mathbf{g}(\mathbf{x}_k, \mathbf{c}_1, \mathbf{c}_2)$. The true position could be anywhere inside the search zone – circle with radius $d_{\max}(k)$ - centered at $\mathbf{g}(\mathbf{x}_k, \mathbf{c}_1, \mathbf{c}_2)$. Initially, the value of the radius is $d_{\max}(k) = d_{\max 0}$ for all $k$. Recall, that $\mathbf{c}_1$ and $\mathbf{c}_2$ are initialized based on longitude and latitude of the four corners of the reference and template images as described in subsection 3.2.

We propose to perform the search of PCs within the search zone w.r.t. only the translation vector components in both directions keeping rotation angle and scaling factor values fixed at their initial values, $\alpha_{Init}$ and $\Delta r_{Init}$. To justify this, let us evaluate the error $d_{10}$ if $\mathbf{A}_{RST}$ matrix is calculated based on the coarse registration matrix $\mathbf{A}_{Init}$: $\mathbf{A}_{RST.Init} = \Delta r_{Init} \begin{pmatrix} \cos(\alpha_{Init}) & \sin(\alpha_{Init}) \\ -\sin(\alpha_{Init}) & \cos(\alpha_{Init}) \end{pmatrix}$. In this case, $d_{10}$ varies from 0.05 to 0.7 pixels. This distortion is symmetrical w.r.t. RI fragment center and will not cause bias of estimation of PCs position. Its main effect will be a slightly reduced correlation between RI and TI fragments. At this stage of our research, we have neglected this error. Note that an iterative refinement of PCs positions using more and more accurate RST model parameters obtained in the registration process can help compensating this error.

We stress that the search w.r.t. translations does not mean that the RAE method cannot be applied to images with higher degree of non-linearity. In this case, additional search of PCs w.r.t. rotation angle and scaling factor is to be considered at the expense of higher computational complexity of PCs search procedure. This does not also mean that we deal with global translation estimation because estimated translations are different for each CF and are estimated independently.

Accordingly, each template CF is projected into the reference image using initial rotation angle

$\alpha_{Init}$ and the scaling factor $\Delta r_{Init}$ and interpolated to the reference image grid so as to form a CF of size $N_{RI}$ (pixels having no match at the template CF are filled with zero values and are no more used in subsequent processing). Similarity between a template CF and a reference CF is measured by NCC in this paper. Within the search zone, NCC may exhibit a multi-extremal behavior with one extremum relating to the true correspondence (if such a correspondence exists for a particular CF) and a number of local extrema relating to false correspondences. The pairs of reference and template CFs linked to each NCC local extremum will be later referred to as putative correspondences, PCs.

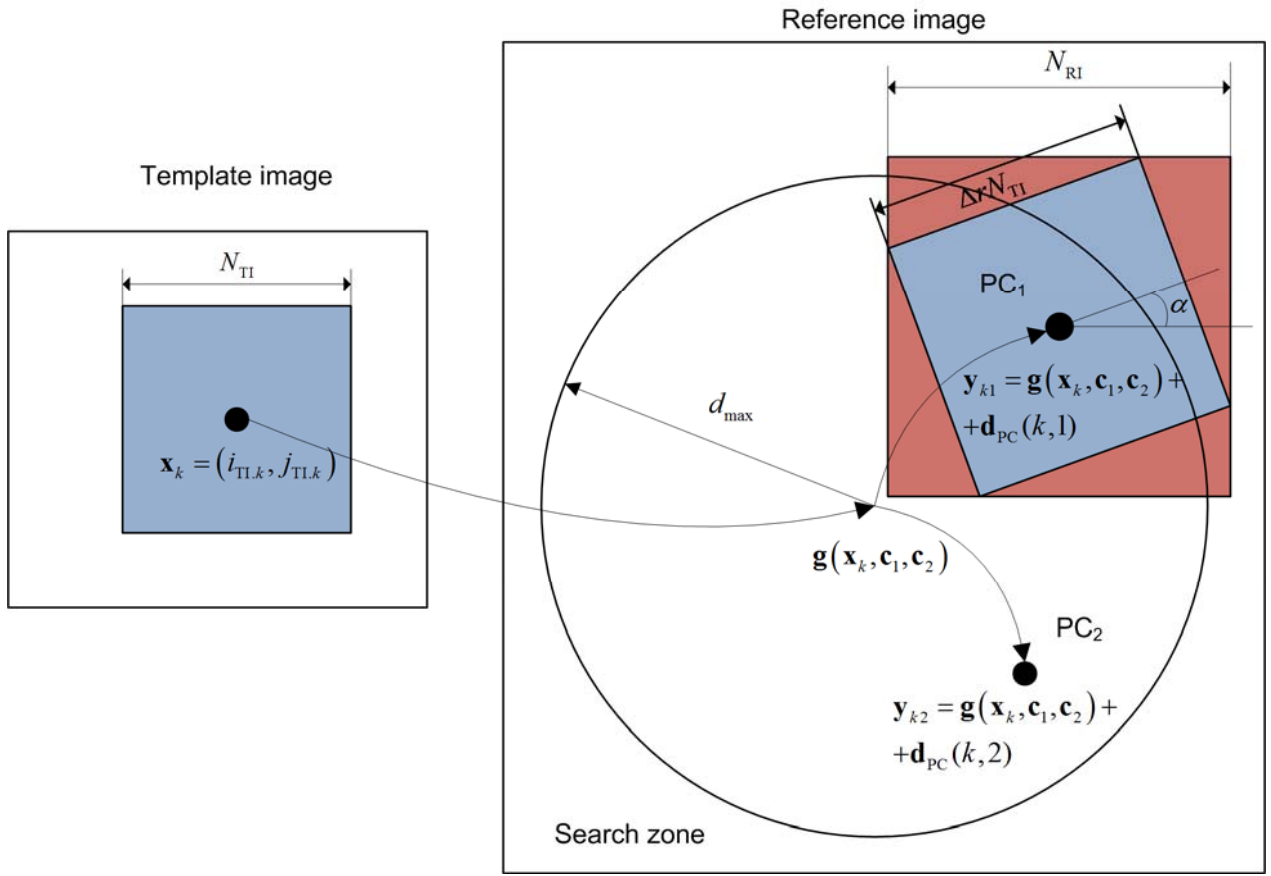

Fig. 1. Illustration of search of putative correspondences between template and reference images

Putative correspondence search is implemented as a two-step procedure: detection of PCs and refinement of their position with subpixel accuracy. First, normalized correlation coefficient, $k_{RT}$, between transformed template CF and all reference CFs in the search zone is calculated on the grid with the lag equal to 0.5 pixel in both directions. This particular value for the lag was found experimentally as a good compromise between computational complexity (that increases with lag decrease), and probability of finding local NCC maxima (that decreases with lag increase). Then, all local NCC extrema with $|k_{RT}| > 0.25$ are found. For a $k$-th template CF, a $p$-th PC is described by the pair of coordinates $\mathbf{x}_k$ and $\mathbf{y}_{kp}$ and the NCC value $k_{RT.kp}$. Second, positions of found PCs are refined with subpixel accuracy. This is done using template image intensity interpolation allowing calculation of NCC for arbitrary RST parameters [39]. NCC value is then maximized w.r.t. $\mathbf{y}_{kp}$.

### 3.4. The proposed method general structure

Modern sensors tend to acquire images of a very large size. Hence, the number of control fragments and putative correspondences can be huge. Therefore, we propose an iterative registration scheme called RAE (Fig. 2) with successive refinement of estimates of geometrical transformation parameters and removal of false PCs.

At the initialization stage, the list of PCs is populated using NCC values. All found PCs are sorted according to the $|k_{RT}|$ value. If they lie within the current search zone, they are considered as active. Then, each iteration proceeds as follows: for the first unprocessed active PCs, potential registration accuracy between the reference and template CFs (in the form of Cramér–Rao Lower Bound, CRLB, on translation estimation error) is estimated. If the registration accuracy is high enough (translation estimation error is low enough), this PC is validated for being involved in the geometrical transformation parameter estimation; otherwise, it is discarded. When a predefined number of validated PCs is found, the estimates of geometrical transform parameter vectors $\mathbf{c}_1$ and $\mathbf{c}_2$ are refined along with their estimation accuracy. Refined

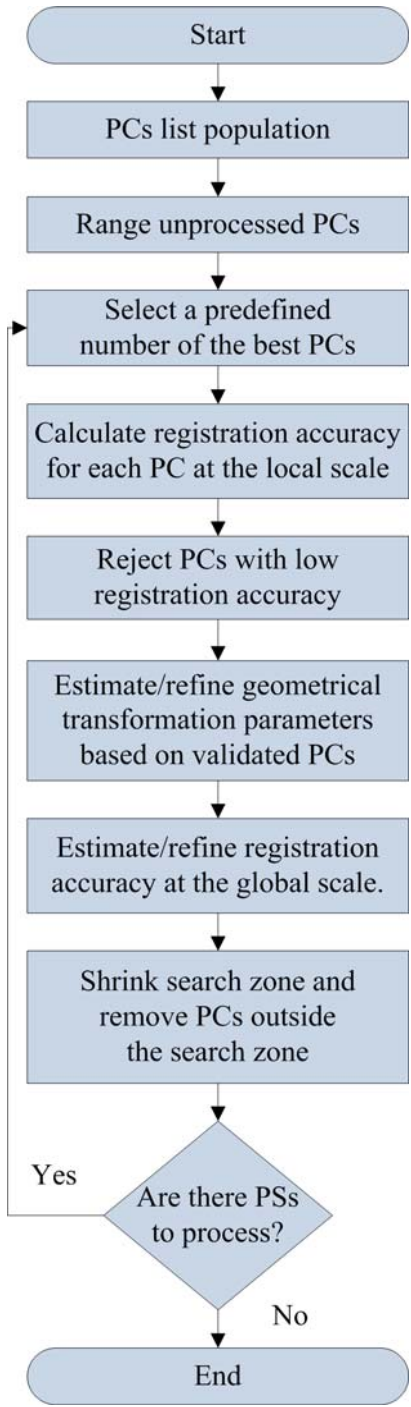

Fig.2. Flow-chart of the proposed RAE registration method

geometrical transform allows reducing the search zone and it shrinks iteratively the list of active PCs. Thus, the computational complexity of each next iteration significantly reduces. Iterations continue until no active PC is left.

The core feature of the proposed RAE registration algorithm is its ability to quantify the registration accuracy. This is done in two ways: first, at local scale, by systematically estimating registration accuracy for each PC, and, second, at global scale, by quantifying estimation accuracy of the transformation parameters $\mathbf{c}_1$ and $\mathbf{c}_2$. Let us next address these stages more in detail.

**3.5. Registration accuracy at the local scale**

At the local scale, for a given PC, we distinguish two factors influencing the registration error. The first one is straightforwardly related to the inherent structure of the registered reference and template fragments and establishes the lowest possible or potential registration accuracy. This factor is described by the potential translation estimation error standard deviation $\sigma_{PC.LB}$. The value of $\sigma_{PC.LB}$ is different for each PC and does not depend on the used registration algorithm. The second factor is the statistical efficiency [40] of the used registration method denoted as $e_{est}$. Recall that $0 \leq e_{est} \leq 1$. When $e_{est} = 1$, the estimator is called efficient and it provides the highest possible registration accuracy equal to $\sigma_{PC.LB}$. In practice, the registration accuracy is lower in some degree due to finite sample size or implementation peculiarities intended for speeding-up. Naturally, higher values of $e_{est}$ are preferable. Though the value $e_{est}$ may depend on the PC (registration method could be more effective for one texture and less effective for another one), such subtle effects are outside the scope of this paper. Following the previous study of NCC estimator behavior in [20, 26, 27], the value of $e_{est}$ is set in all experiments equal to 0.1.

Taking into account these two factors, for each PC, the registration accuracy is calculated as

$$\sigma_{PC} = \sigma_{PC.LB} / \sqrt{e_{est}} . \qquad (3)$$

To estimate $\sigma_{PC.LB}$, we rely on our previous works [20, 26, 27] where CRLB on RST parameter estimation error was obtained and studied. For each PC, calculation of $\sigma_{PC.LB}$ involves the following stages. At the initialization stage, noise parameters are defined separately for reference and template fragments. We use complex model applicable to RS images of different types: spatially correlated normally distributed noise with signal-dependent variance [27]. Noise model parameters are estimated in advance according to a procedure specified in Section 4. At the first stage, PC texture parameter vector $\boldsymbol{\theta}_{\text{PC.texture}} = (\sigma_{x.\text{RI}}, \sigma_{x.\text{TI}}, k_{\text{RT}}, H)$ is estimated. Here $\sigma_{x.\text{RI}}, \sigma_{x.\text{TI}}$ define standard deviation of texture increments on unit distance for reference and template CFs, and Hurst exponent $H \in (0,1)$ characterizes texture roughness (value less than 0.5 corresponds to rough and greater than 0.5 - to smooth textures). The estimation of the value of $k_{\text{RT}}$ is improved at this stage thanks to the more relevant noise model used. At the second stage, the estimated texture parameter vector is appended by translation estimates $\mathbf{d}_{\text{PC}} = \mathbf{y}_{kp} - \mathbf{g}(\mathbf{x}_k, \mathbf{c}_1, \mathbf{c}_2)$ to form 6 by 1 vector $\boldsymbol{\theta}_{\text{PC}} = (\hat{\sigma}_{x.\text{RI}}, \hat{\sigma}_{x.\text{TI}}, \hat{k}_{\text{RT}}, \hat{H}, \mathbf{d}_{\text{PC}}(1), \mathbf{d}_{\text{PC}}(2))$. Rotation angle and scaling factor values are the same as for the PCs search stage and they are not included in $\boldsymbol{\theta}_{\text{PC}}$ as being considered fixed. The corresponding Fisher Information Matrix (FIM) $\mathbf{I}_{\boldsymbol{\theta}_{\text{PC}}}$ on $\boldsymbol{\theta}_{\text{PC}}$ vector is then evaluated and inverted to obtain CRLB matrix $\mathbf{C}_{\boldsymbol{\theta}_{\text{PC}}} = \mathbf{I}_{\boldsymbol{\theta}_{\text{PC}}}^{-1}$. The submatrix $\mathbf{C}_{\boldsymbol{\theta}_{\text{PC}}}(5:6, 5:6)$ characterizes potential translation estimation accuracy or PC registration accuracy. As it was shown in [20], the estimation errors of the two translation components might be correlated and have different standard deviations. But these effects are not essential and $\mathbf{C}_{\boldsymbol{\theta}_{\text{PC}}}(5:6, 5:6)$ can be well approximated as scalar multiple of an identity matrix. Therefore, $\sigma_{\text{PC.LB}}$ is finally obtained as $\sigma_{\text{PC.LB}} = \sqrt{(\mathbf{C}_{\boldsymbol{\theta}_{\text{PC}}}(5,5) + \mathbf{C}_{\boldsymbol{\theta}_{\text{PC}}}(6,6))/2}$.

For each $kp$-th PC, the estimated value $\sigma_{\text{PC.LB}.kp}$ is compared to the predefined threshold $\sigma_{\text{PC.LB.max}}$. If $\sigma_{\text{PC.LB}.kp} < \sigma_{\text{PC.LB.max}}$, this PC is considered suitable for registration with respective registration accuracy $\sigma_{\text{PC}.kp} = \sigma_{\text{PC.LB}.kp} / \sqrt{e_{\text{est}}}$. Such a PC is called CRLB-validated PC and denoted further as vPC. The number of vPCs for a $k$-th CF is denoted as $n_{\text{vPC}}(k)$; their total number is $n_{\text{vPC}} = \sum_{k=1}^{N_{CF}} n_{\text{vPC}}(k)$. In the experimental part of the paper, we set $\sigma_{\text{PC.LB.max}} = 0.35$ pixel for all test cases. This choice is based on the analysis performed in [27] where it was shown that registration errors using NCC similarity measure were linearly related to $\sigma_{PC.LB}$ up to values $\sigma_{PC.LB} < 0.4$. For higher values of $\sigma_{PC.LB}$ (related either to higher noise level or lower correlation between reference and template images), NCC registration becomes unreliable.

**3.6. Geometrical transformation parameters estimation. Registration accuracy at global scale**

To detect true correspondences and to estimate more reliably the transformation parameters, we follow the likelihood approach developed in [22-24]. Let us first describe the proposed solution and subsequently discuss its distinctive features and advantages.

At the global scale, we process only CFs with at least one CRLB-validated PC (CF with $n_{vPC}(k) > 0$). Their number is $N_{vCF}$. For notation simplicity, we use the same indexes $k$ and $p$ for CRLB-validated CFs and PCs.

For each $k$-th CF, we define a binary valued vector $\mathbf{z}_k$ with $n_{vPC}(k)$ elements. Unity in $p$-th position of the vector $\mathbf{z}_k$ indicates that $p$-th PC is an inlier. In this case, probability density function (p.d.f.) of observing $\mathbf{y}_{kp}$ value is $N\left(\mathbf{y}_{kp} - \mathbf{g}(\mathbf{x}_k, \mathbf{c}_1, \mathbf{c}_2), \mathbf{\Sigma}_{kp}\right)$, where $\mathbf{\Sigma}_{kp} = \begin{pmatrix} \sigma_{PC.kp}^2 & 0 \\ 0 & \sigma_{PC.kp}^2 \end{pmatrix}$ is the covariation matrix of registration error. Zeros in a $p$-th position of the vector $\mathbf{z}_k$ indicate that the $p$-th PC is an outlier. In this case, $\mathbf{y}_{kp}$ is distributed uniformly within the respective search zone with p.d.f. $1/S_{SearchZone}(k)$, where $S_{SearchZone} = \pi d_{\max}^2(k)$ is the area of the search zone. Probability of a $p$-th PC to be an inlier is defined as $P(z_k(p) = 1) = P_{kp}^{in}$. Correspondingly, $P(z_k(p) = 0) = 1 - P_{kp}^{in}$. Since only one among PCs related to the same CF can be an inlier, only one element of $\mathbf{z}_k$ or none can be unity.

Events when the 1st, 2nd,…, $k$ th,…, or $n_{vPC}(k)$ PC is a true correspondence are disjoint and equiprobable with *a priori* probability $P_{PC}^{in}$. The value of $P_{PC}^{in}$ is derived using *a priori* probability $P_{CF}^{in}$ to find a true correspondence for a given CF as $P_{PC}^{in}(k) = 1 - \left(1 - P_{CF}^{in}\right)^{1/n_{vPC}(k)}$. Note that $P_{CF}^{in}$ is considered as a fixed value for all CFs while $P_{PC}^{in}$ varies from CF to CF depending on $n_{vPC}(k)$. The p.d.f. of PCs related to a single CF conditional on $\boldsymbol{\theta} = \left( P_{CF}^{in}, \mathbf{c}_1, \mathbf{c}_2 \right)$ is

$$f\left(\mathbf{y}_{k1}, \mathbf{y}_{k2}, \ldots, \mathbf{y}_{kn_{vPC}(k)} / \mathbf{x}_k, \boldsymbol{\theta}\right) = \sum_{\mathbf{z}} f\left(\mathbf{y}_{k1}, \mathbf{y}_{k2}, \ldots, \mathbf{y}_{kn_{vPC}(k)}, \mathbf{z} / \mathbf{x}_k, \boldsymbol{\theta}\right) = $$
$$= P_{PC}^{in}(k) \sum_{p=1}^{n_{vPC}(k)} N\left(\mathbf{y}_k - \mathbf{g}(\mathbf{x}_k, \mathbf{c}_1, \mathbf{c}_2), \boldsymbol{\Sigma}_{kp}\right) + \left(1 - P_{CF}^{in}\right) / S_{SearchZone}(k). \qquad (4)$$

Using the independence between CFs, the p.d.f. for all CFs is obtained as:

$$f\left(\mathbf{y}_{11}, \mathbf{y}_{12}, \ldots, \mathbf{y}_{N_{vCF} n_{vPC}(N_{CF})} / \mathbf{x}_1, \mathbf{x}_2, \ldots \mathbf{x}_{N_{vCF}}, \boldsymbol{\theta}\right) = \prod_{k=1}^{N_{vCF}} f\left(\mathbf{y}_{k1}, \mathbf{y}_{k2}, \ldots, \mathbf{y}_{kn_{vPC}(k)} / \mathbf{x}_k, \boldsymbol{\theta}\right). \qquad (5)$$

The p.d.f. (5) has one drawback that we will illustrate with an example. Fig. 3 shows a specific composition of CFs where all but one are in one cluster whilst one CF stands aside. For the clustered CFs, the estimates $\hat{\mathbf{c}}_1$ and $\hat{\mathbf{c}}_2$ will average the influence of all CFs in the cluster: none CF has a preponderant role. For the isolated CF, flexibility of the transformation $\mathbf{g}(\mathbf{x}, \mathbf{c}_1, \mathbf{c}_2)$ allows approximating large difference between $\mathbf{y}_{kp}$ and $\mathbf{g}(\mathbf{x}_k, \mathbf{c}_1, \mathbf{c}_2)$ making $\mathbf{y}_{kp} - \mathbf{g}(\mathbf{x}_k, \hat{\mathbf{c}}_1, \hat{\mathbf{c}}_2) \approx 0$ at the same time not altering the cluster registration. Therefore, probability of the isolated PC to be an inlier defined by p.d.f. $N\left(\mathbf{y}_{kp} - \mathbf{g}(\mathbf{x}_k, \mathbf{c}_1, \mathbf{c}_2), \boldsymbol{\Sigma}_{kp}\right)$ will be high irrespectively of whether this point is an inlier or an outlier. This effect is called leverage in statistics (particularly in regression analysis) [28], where isolated points lacking neighboring observations are called high-leverage points. Notice that researchers in the image registration area are also well aware of this problem. The benefit that can be drawn from control fragments clustering has been systematically outlined and used for improving the registration accuracy [41, 42].

We propose to solve this problem directly by introducing a modification to p.d.f. (4) according to the leave-one-out cross-validation method [12]. The idea is to define p.d.f. of observing the inlying PC $\mathbf{y}_{kp}$ as $N\left(\mathbf{y}_{kp} - \mathbf{g}(\mathbf{x}_k, \mathbf{c}_{1k}, \mathbf{c}_{2k}), \boldsymbol{\Sigma}_{kp}\right)$, where $\mathbf{c}_{1k}$ and $\mathbf{c}_{2k}$ are transformation parameters defined by all PCs excluding ones related to the $k$ th CF. In this manner, $\mathbf{g}(\mathbf{x}_k, \hat{\mathbf{c}}_{1k}, \hat{\mathbf{c}}_{2k})$ is a prediction of $\mathbf{y}_{kp}$ based on PCs related to the neighboring CFs. Such a prediction has low probability to be accurate for outliers and excludes convergence of $\mathbf{y}_{kp} - \mathbf{g}(\mathbf{x}_k, \hat{\mathbf{c}}_1, \hat{\mathbf{c}}_2)$ to small values for isolated CFs. In statistical terms, our score function assures absence of high-leverage points (or points with high Cook's distance [28] that is also based on leave-one-out cross-validation method) among inliers.

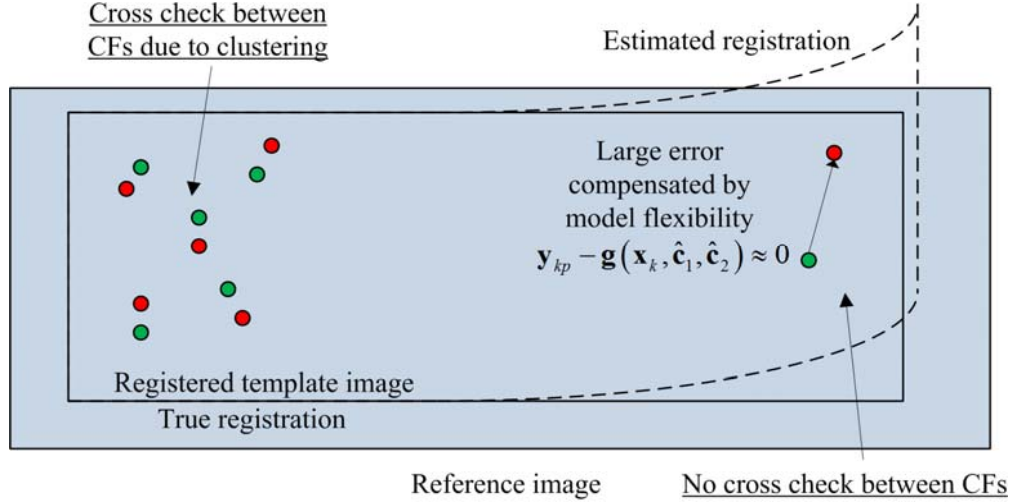

Fig.3. Illustration of CFs clustering effect

The corrected p.d.f. of PCs related to a single CF conditional on $\boldsymbol{\theta}_k = \left(P_{CF}^{in}, \mathbf{c}_{1k}, \mathbf{c}_{2k}\right)$ becomes

$$f\left(\mathbf{y}_{k1}, \mathbf{y}_{k2}, \ldots, \mathbf{y}_{kn_{\text{vPC}}(k)} / \mathbf{x}_k, \boldsymbol{\theta}_k\right) = P_{PC}^{in}(k) \sum_{p=1}^{n_{\text{vPC}}(k)} N\left(\mathbf{y}_k - \mathbf{g}(\mathbf{x}_k, \mathbf{c}_{1k}, \mathbf{c}_{2k}), \boldsymbol{\Sigma}_{kp}\right) + \left(1 - P_{CF}^{in}\right) / S_{SearchZone}(k), \quad (6)$$

and p.d.f. for all CFs is now given by:

$$f\left(\mathbf{y}_{11}, \mathbf{y}_{12}, \ldots, \mathbf{y}_{N_{vCF} n_{\text{vPC}}(N_{CF})} / \mathbf{x}_1, \mathbf{x}_2, \ldots \mathbf{x}_{N_{vCF}}, \boldsymbol{\theta}\right) = \prod_{k=1}^{N_{CF}} f\left(\mathbf{y}_{k1}, \mathbf{y}_{k2}, \ldots, \mathbf{y}_{kn_{\text{vPC}}(k)} / \mathbf{x}_k, \boldsymbol{\theta}_k\right) \quad (7)$$

where $\boldsymbol{\theta} = \left(P_{CF}^{in}, \mathbf{c}_{11}, \mathbf{c}_{21}, \ldots, \mathbf{c}_{1N_{vCF}}, \mathbf{c}_{2N_{vCF}}\right)$. Note that the parameter vector $\boldsymbol{\theta}$ now includes the pairs $\mathbf{c}_{1k}$ and $\mathbf{c}_{2k}$ for all $N_{vCF}$ control fragments.

The problem (7) is solved via EM-algorithm. The complete-data log-likelihood takes the following form:

$$Q(\boldsymbol{\theta}, \hat{\boldsymbol{\theta}}) = \sum_{k=1}^{N_{vCF}} \sum_{p=1}^{n_{\text{vPC}}(k)} \left[ -\frac{P_{kp}^{in}(\hat{\boldsymbol{\theta}})}{2\sigma_{PC.kp}^2} \left\|\mathbf{y}_{kp} - \mathbf{g}(\mathbf{x}_k, \mathbf{c}_{1k}, \mathbf{c}_{2k})\right\|^2 - \ln(\sigma_{PC.kp}^2) P_{kp}^{in}(\hat{\boldsymbol{\theta}}) + \right.$$
$$\left. + \ln\left(P_{PC}^{in}(k)\right) P_{kp}^{in}(\hat{\boldsymbol{\theta}}) + \ln\left(1 - P_{CF}^{in}\right)\left(1 - P_{kp}^{in}(\hat{\boldsymbol{\theta}})\right) + \ln\left(1 / S_{SearchZone}(k)\right) \right]. \quad (8)$$

At the E-step of EM-method, the probabilities $P_{kp}^{in}(\hat{\boldsymbol{\theta}})$, i.e. *a posteriori* probability for each correspondence to be an inlier, are estimated. Using Bayes' rule, we get

$$\hat{P}_{kp}^{in} = \frac{P_{PC}^{in}(k)N(\mathbf{y}_{kp} - \mathbf{g}(\mathbf{x}_k, \hat{\mathbf{c}}_{1k}, \hat{\mathbf{c}}_{2k}), \mathbf{\Sigma}_{kp})}{P_{PC}^{in}(k)N(\mathbf{y}_{kp} - \mathbf{g}(\mathbf{x}_k, \hat{\mathbf{c}}_{1k}, \hat{\mathbf{c}}_{2k}), \mathbf{\Sigma}_{kp}) + (1 - \hat{P}_{CF}^{in})/S_{SearchZone}(k)}. \quad (9)$$

The M-step of the algorithm is responsible for maximization of $Q(\boldsymbol{\theta}, \hat{\boldsymbol{\theta}})$ with respect to $\boldsymbol{\theta}$. The average probability to find a true correspondence for a single CF is found as

$$\hat{P}_{CF}^{in} = \frac{1}{N_{vCF}} \sum_{k=1}^{N_{vCF}} \hat{P}_{CF}^{in}(k),$$

where $\hat{P}_{CF}^{in}(k) = \left[1 - \prod_{p=1}^{n_{vPC}(k)}(1 - \hat{P}_{kp}^{in})\right]$ is *a posteriori* probability that at least one of PCs of $k$-th CF is an inlier. The inlying PCs are found at this stage according to the rule $\hat{P}_{kp}^{in} > P_{th}^{in}$, where $P_{th}^{in}$ is a threshold. If more than one PC belonging to the same CF is recognized as an inlier, the one with maximal $\hat{P}_{kp}^{in}$ is selected. CFs with inlying PC are called inlying CFs. In the experiments, we set $P_{th}^{in} = 0.9$ for all test cases.

Maximization of (8) with respect to $\mathbf{c}_{1k}$ and $\mathbf{c}_{2k}$ is a weighted polynomial regression problem with weights of each PC inversely proportional to $w_{kp} = \sigma_{PC.kp}^2 / \hat{P}_{kp}^{in}$. Solution of this problem is found by equating the first derivatives of (8) to zero with respect to elements of $\mathbf{c}_{1k}$ and $\mathbf{c}_{2k}$:

$$\hat{\mathbf{c}}_{1k} = \mathbf{I}_k^{-1} \cdot \mathbf{b}_{1k} \quad \hat{\mathbf{c}}_{2k} = \mathbf{I}_k^{-1} \cdot \mathbf{b}_{2k}, \quad (10)$$

where $\mathbf{I}_k = \sum_{t \neq k} \sum_{p=1}^{n_{vPC}(t)} w_{tp}^{-1} \mathbf{e}_{tp}^T \mathbf{e}_{tp}$, $\mathbf{b}_{1k} = \sum_{t \neq k} i_{RI.t} \sum_{p=1}^{n_{vPC}(t)} w_{tp}^{-1} \mathbf{e}_{tp}^T$, $\mathbf{b}_{2k} = \sum_{t \neq k} j_{RI.t} \sum_{p=1}^{n_{vPC}(t)} w_{tp}^{-1} \mathbf{e}_{tp}^T$,

$\mathbf{e}_{tp} = (1, i_{TI.tp}, j_{TI.tp}, ..., i_{TI.tp}^{k_1} j_{TI.tp}^{k_2}, ..., i_{TI.tp}^n, j_{TI.tp}^n)$. The first sum is over all inlying CFs except the $k$-th CF. Maximization step fails if the number of inlying CFs is not large enough to solve linear equations in (10), that is, less than $n_c + 1$.

The complexity of (10) for one CF is $O(n_{vPC}) = O(N_{vCF}) = O(N_{CF})$; for all CFs - $O(N_{CF}^2)$. Quadratic complexity is undesirable. Therefore, we propose to estimate $\mathbf{c}_{1k}$ and $\mathbf{c}_{2k}$ using a fixed number of inlying CFs instead of all CFs. We set this number to the minimum sufficient value $n_c + 1$. Using k-d trees [43], the complexity of search for a finite number of closest neighbors is $O(N_{CF} \ln(N_{CF}))$. As a result, complexity of global transformation parameter estimation procedure is linearithmic.

The final geometrical transformation parameters $\hat{\mathbf{c}}_1$ and $\hat{\mathbf{c}}_2$ are estimated after the EM-algorithm convergence according to the same formula (10) but using all validated CFs. In this case, the matrix $\mathbf{I}_k$ is replaced by $\mathbf{I}$ in (10). The estimation errors of $\hat{\mathbf{c}}_1$ and $\hat{\mathbf{c}}_2$ are independent of each other and characterized by the same covariation matrix $\mathbf{R}_c$ equal to the $\mathbf{I}^{-1}$ (as $\mathbf{I}$ matrix is, in essence, a Fisher Information Matrix).

The matrix $\mathbf{R}_c$ characterizes registration accuracy of the proposed method at the global scale. Note that it is a function of registration accuracy at the local scale $\sigma^2_{PC.kp}$. Given covariation matrix $\mathbf{R}_c$, the registration accuracy at point $(i_{TI}, j_{TI})$ at TI image is obtained as:

$$\sigma_{reg}(i_{TI}, j_{TI}) = \sqrt{\left[1, i_{TI}, j_{TI}, ..., i_{TI}^{k_1} j_{TI}^{k_2}, ..., i_{TI}^{n} j_{TI}^{n}\right] \mathbf{R}_c \left[1, i_{TI}, j_{TI}, ..., i_{TI}^{k_1} j_{TI}^{k_2}, ..., i_{TI}^{n} j_{TI}^{n}\right]^T}, \quad 0 \le k_1 + k_2 \le n.$$

Note that $\sigma_{reg}(i_{TI}, j_{TI})$ is in RI pixels. The respective point at RI image is defined by estimated parameter vectors $\hat{\mathbf{c}}_1$ and $\hat{\mathbf{c}}_2$ according to model (2). Later we use the notion $\sigma_{reg}(k) = \sigma_{reg}(i_{TI.k}, j_{TI.k})$ to describe the registration accuracy for the $k$-th CF.

One drawback of the EM-algorithm is that it does not assure finding the global maximum of a likelihood function [44]. It might converge to a local maximum depending on an initial guess for global transformation parameters. To increase the probability of finding the global maximum, we propose to use a multistart approach [45]. As discussed above, RS sensors are highly linear (neglecting relief influence) and an affine transformation as an initial guess is a reasonable choice.

Multistart optimization runs as follows. Random triples of vPCs belonging to different CFs are selected without repetition. For each triple, affine transformation parameters are estimated. The found transformation is considered as valid if the maximum difference between initial and newly found transformation over the reference image area does not exceed $d_{\max 0}$. Each valid transformation is used as initial guess for the EM-algorithm. The algorithm stops when a fixed number of validated initial guesses is found. The solution that maximizes the complete-data log-likelihood is then chosen. We have found experimentally that 10 starts are sufficient.

Let us summarize below the key differences between the proposed registration method and the method [23] published recently:

(1) Multiple putative correspondences for each CF are taken into account;

(2) The leave-one-out cross-validation approach is utilized, the registration method is formulated purely as a likelihood maximization without introducing any additional empirical regularization terms like Local Linear Transformation term in [23]. Owing to this, the RAE method is less restrictive with respect to geometrical transformation properties.

(3) Registration accuracy for each putative correspondence (in the form of CRLB) is introduced into the likelihood function as additional *a priori* information. By doing this, we take into account the structural differences between PCs affecting registration accuracy.

The most important consequence of these modifications is that it becomes possible to obtain registration accuracy (variance) at the global scale. In the experimental part of the paper, we will demonstrate that the registration accuracy estimated in this manner is very accurate even when only a few correspondences are available.

**3.7. Reduction of computational complexity**

Processing of a single PC according to the proposed method is a computationally intensive task (though all processing stages except for EM optimization stage are of linear complexity w.r.t number of CFs). First of all, CRLB calculation involves operations with joint correlation matrix of reference and template image fragments. Second, the proposed method emphasizes registration accuracy, therefore, time consuming area-based methods with subpixel registration accuracy are needed for PC search and registration (like NCC). To cope with the arising complexity problem, below we propose an implementation of the RAE method allowing to drastically reduce the number of processed PCs making the registration complexity acceptable.

The basic idea under speeding up the proposed method is that each newly found CRLB-verified PC can, in principle, contribute to reduce uncertainty about global transformation parameters. The refined estimation of global transformation parameters, in turn, allows reducing search zone and removing PCs outside the shrunk search zone. Such an alternative scheme requires precise prediction of registration accuracy at each stage - this is the main feature of the proposed method.

The RAE method is detailed in Alg. 1. It includes the population stage of PCs list (2-3), PC processing stage (5-6), global transformation parameters refinement stage (8) and PCs list truncation stage (9). Finding all PCs in advance may be unreasonable as some of them will be rejected later at the global transformation refinement stage. Therefore, we balance time between PC list population and CRLB calculation (stage 4). In this manner, the PC population list becomes asynchronous. PCs are processed starting from the one with the highest $\left|k_{\text{RT}.kp}\right|$ value. At each $t$-th iteration, PC processing stage runs until a predefined number of CRLB-validated PC, $n_{\text{PCnew}}(t)$, is found. After this, the estimate of geometrical transformation parameters is refined, the registration SD $\sigma_{reg}(k)$ for each CFs is calculated and used to shrink the search zone according to the expression $d_{\max}(k) = 6 \cdot \sigma_{reg}(k) + 2$. Here, $d_{\max}(k)$ is the 6-sigma zone enlarged by 2 pixels to account for the width of NCC lobes (even for perfectly registered images, PCs identified by the NCC measure deviate more or less from the true correspondence). We used 6-sigma interval instead of 3-sigma interval to account for experimental results indicating that $\sigma_{reg}(k)$ underestimates real RAE method performance. Finally, the PC list is truncated by rejecting all PCs outsize the newly defined search zone.

As it was shown, the global transform parameter estimation procedure is of $O\left(\ln(N_{CF})N_{CF}\right)$ complexity. Iterative nature of the proposed registration method suggests recalculation of the global transform parameters after each $n_{\text{PCnew}}(t)$ processed CFs and gradual reduction of the search zone with respect to spatial coordinates.

Algorithm 1: RAE registration

**Input**: Reference and template images, product corner coordinates, direct geopositioning errors;

**Output**: Coefficients of polynomial geometrical transform model, covariation matrix of the coefficients estimation error;

1. Set $t = 1$, set initial value $n_{PCnew}(1)$;
2. Randomly select one template CF among unprocessed ones;
3. Find PC for the selected CF and populate the PCs list;
4. Repeat 2-3 until the processing time exceeds time for calculating $CRLB_{fBm}$ for one PC. Skip the stages 2-3 if all CFs are processed (the list of PCs has been populated);
5. Select PC with the maximum value of $|k_{RT.kp}|$;
6. Calculate CRLB $\sigma_{PC.kp}$ on registration error;
7. Repeat the steps from 2 to 6 until new $n_{PCnew}(t)$ CRLB-validated PCs are found;
8. Estimate the global transform parameters. Evaluate the registration accuracy $\sigma_{reg}(k)$ of each CF;
9. Determine the new search zone for each CF as $d_{max}(k) = \min\left(6 \cdot \sigma_{reg}(k) + 2, d_{max.0}\right)$, remove PCs outside the newly calculated search zone;
10. Set $t = t + 1$. Calculate $n_{PCnew}(t)$;
11. Repeat the steps from 2 to 10 until all PCs are processed.

Let us first consider a fixed step size $n_{PCnew}(t) = n_{PCnew}$ and define the expected number of iterations as $r_{const} = n_{vPC} / n_{PCnew}$, where $n_{vPC}$ is the total number of CRLB-validated PC. In this case, complexity for determining the expected number of iterations $O(r_{const}) = O(n_{vPC}) = O(N_{CF})$ and the total complexity of the global transform parameters estimation procedure exceed quadratic:

$$O\left[n_{PCnew} \ln(n_{PCnew}) + 2n_{PCnew} \ln(2n_{PCnew}) + ... + r_{const} n_{PCnew} \ln(r_{const} n_{PCnew})\right] =$$

$$= O\left[\ln(n_{PCnew}) n_{PCnew} (1 + 2 + ... + r_{const}) + n_{PCnew} (1 + 2\ln(2) + ... + r_{const} \ln(r_{const}))\right] =$$

$$= O\left(r_{\text{const}}^2 \ln(r_{\text{const}})\right) = O\left(N_{\text{SW}}^2 \log(N_{\text{SW}})\right).$$

Frequent recalculation of the global transform parameters is needed only at the beginning of the registration process when each newly found correspondence can significantly improve the overall registration accuracy. Therefore, as a second and more effective strategy, we propose linear increase of $n_{\text{PCnew}}(t)$ step size:

$$n_{\text{PCnew}}(t+1) = q n_{\text{PCnew}}(t), \; n_{\text{PCnew}}(1) = 1.$$

The last iteration index is $r_{\text{linear}} = \ln(n_{\nu PC}) / \ln(q)$. Note that $O(r_{\text{linear}}) = O(\ln(n_{\nu PC})) = O(\ln(N_{\text{CF}}))$ and $O(q^{r_{\text{linear}}}) = O(n_{\nu PC}) = O(N_{\text{CF}})$. In this case, complexity of the global transform parameters estimation remains linearithmic:

$$O\left(q \ln(q) + q^2 \ln(q^2) + \ldots + q^{r_{\text{linear}}} \ln(q^{r_{\text{linear}}})\right) = O\left(\ln(q)\left(q + 2q^2 + \ldots + r_{\text{linear}} q^{r_{\text{linear}}}\right)\right) =$$

$$O(r_{\text{linear}} q^{r_{\text{linear}}}) = O\left(N_{\text{SW}} \ln(N_{\text{SW}})\right).$$

Thus, the overall linearithmic complexity of the RAE method w.r.t. number of CFs is assured.

## 4. EXPERIMENTAL PART

**4.1. Test data description**

Eight registration cases are considered to analyze capabilities of the proposed RAE method: (1, 5) optical-to-optical, (2, 6) optical-to-DEM, (3, 7) optical-to-radar, and (4, 8) DEM-to-radar image pairs. A detailed description of each case is given in Table 1. The first and fifth cases relate to monomodal registration, others relate to complex multimodal registration. All cases correspond to multitemporal framework with differences in acquisition time from 7 to 22 years.

The test cases 1 and 2 share the same band of the reference Hyperion image, the test cases 3 and 4 exploit the same SIR-C template image (HH polarization channel). These images after registration are shown in Fig. 4 and 5. These images and data in Table 1 reveal the complexity for each test case. Test images for the test cases 5…8 are not shown for avoiding a too lengthy paper. The used test images exhibit a wide variety of land covers: urban, rural, forest, agricultural, rivers, and snow cover. They have very scarce water coverage; no complex water-land boundaries are present except for test case 6. It is well known that the presence of such boundaries facilitates registration and *vice versa*. So, we prefer to deal with complicated practical situations.

Table 1. Characteristics of the test datasets

| Case | Image modality reference template | Sensor/dataset | Acquisition date | Site Latitude/ Longitude, degrees | Spatial resolution, m | Scale | Initial registration error $d_{max0}$, pixels |
|---|---|---|---|---|---|---|---|
| 1 | Optical | Hyperion (band #25) EO1H1800252002116110KZ | 26.04.2002 | 49.4339/ 32.0678 | 30.38 | 1 | 125 |
| | Optical | Landsat8, OLI (band #1) LC81770252014065LGN00 | 06.03.2014 | 48.8497/ 31.6597 | 30 | 1 | |
| 2 | Optical | The same Hyperion band as in case 1 | | | | 1/2 | 130 (65 at scale 0.5) |
| | DEM | ASTER GDEM-2* ASTGTM2_N48-49E031-032 | 2009** | 49/ 32 | 1 arc-second (≈30m at the equator) | 1/2, 1/3 (vertical, (horizontal) | |
| 3 | Optical | Landsat8, OLI (band #8) LC81990262014363LGN00 | 29.12.2014 | 48.8666/ 2.3488 | 15 | 1/2 | 110 (55 at scale 0.5) |
| | Radar | SIR-C (HH polarization) pr41419_ldr_ceos | 05.10.1994 | 48.9584/ 2.8732 | 12.5 | 1/2 | |
| 4 | DEM | ASTER GDEM-2 ASTGTM2_N48-49E002-003 | 2009 | 49/ 3 | 1 arc-second | 1/2, 1/3 | 60 (30 at scale 0.5) |
| | Radar | The same as in case 3 | | | | 1/4 | |
| 5 | Optical | Hyperion (band #155) EO1H2010262006218110PZ | 06.08.2006 | 48.3892/ -1.1613 | 30.38 | 1 | 45 |
| | Optical | Landsat8, OLI (band #5) LC82010262013342LGN00 | 08.12.2013 | 48.8662/ -0.7878 | 30 | 1 | |
| 6 | Optical | The same Landsat8 band as in case 5 | | | | 1/2 | 60 (30 at scale 0.5) |
| | DEM | ASTER GDEM-2 ASTGTM2_N47-48W001-002 | 2009 | 48/ -1 | 1 arc-second | 1/2, 1/3 | |
| 7 | Radar | SIR-C (HH polarization) pr43020_ldr_ceos | 05.10.1994 | 49.8250/ 36.7498 | 12.5 | 1/2 | 120 (60 at scale 0.5) |
| | Optical | Landsat8, OLI (band #8) LC82010262013342LGN00 | 16.02.2016 | 50.2810/ 36.9146 | 15 | 1/2 | |
| 8 | Radar | The same as in case 7 | | | | 1/5 | 60 (30 at scale 1/2) |
| | DEM | ASTER GDEM-2 ASTGTM2_N49-50E036-037 | 2009 | 50/ 37 | 1 arc-second | 1/2, 1/3 | |
| *ASTER GDEM is a product of METI and NASA ; hyphen in the tiles name indicates range of longitudes/lattitudes | | | | | | | |
| **ASTER GDEM-2 release date | | | | | | | |

Coarse registration of the test pairs of images was performed based on metadata information provided with each image: longitudes and latitudes of image corners. For each test case, reference affine transformation between RI and TI was obtained using about 25 manually found control points. Initial registration errors (see Table 1) were measured by comparing coarse and manual registration. For the test cases 1 and 2, initial error in one direction (along-track of the reference Hyperion image) turned out to exceed significantly the assumed interval ±100 pixels and reached values up to 300 pixels. We attribute this, at least, partly, to different reference geodetic datum of Hyperion, Landsat 8, and ASTER GDEM data. To relax this exceedingly high initial error, the corresponding correction shift was introduced: 180 pixels for the test case 1 and 110 pixels for the test case 2. This correction affected only the term $\mathbf{d}_{\text{Initial}}$ in (1) leaving the matrix $\mathbf{A}_{\text{Initial}}$ unchanged, e.g. initial rotation angle and scale factor. Even after this preliminary correction, the initial registration error magnitude exceeded the values expected for RS imagery (±125 pixels for the test case 1, ±130 pixels for the test case 2, and ± 110 pixels for the test case 3). Nevertheless, we kept such an initial error to better illustrate the strength and benefits of the proposed approach.

Images scaling was applied in the test cases 2…4 and 6…8. For cases 2, 4, 6, and 8, the aim was to correct the difference in spatial resolution of DEM images in latitudinal and longitudinal directions. Indeed, ASTER GDEM has the same 1 arc-second spatial resolution in both latitudinal and longitudinal directions, but these resolutions are different when they are expressed in meters. At the considered latitudes, the resolution in latitudinal direction is about 1.5 times higher than in longitudinal direction. Such a difference violates isotropic scaling hypothesis assumed to derive $CRLB_{\text{fBm}}$ bound. That is why, an anisotropic scaling with the factors 1/2 in longitudinal direction and 1/3 in latitudinal direction was applied to the ASTER GDEM data to correct its initial spatial resolution and to make it almost the same in both directions (60m). The optical and radar images in these test cases were rescaled in order to match the DEM spatial resolution. Scaling with the factor 1/2 in the test cases 3 and 7 was applied to take into account the fact that $CRLB_{\text{fBm}}$ bound becomes less adequate at the main scale for SIR-C radar image, as it was shown in [27].

For each image, noise is characterized by spatial correlation function width $\sigma_c$ pixels, and signal-dependent noise variance in the form $\sigma_n^2 = \sigma_a^2 + \sigma_P^2 I + \sigma_\mu^2 I^2$, where $\sigma_a^2$ is additive noise variance, $\sigma_P^2$ is a coefficient defining Poisson component (applicable for optical data) and $\sigma_\mu^2$ is a coefficient responsible for multiplicative noise component (applicable for radar data).

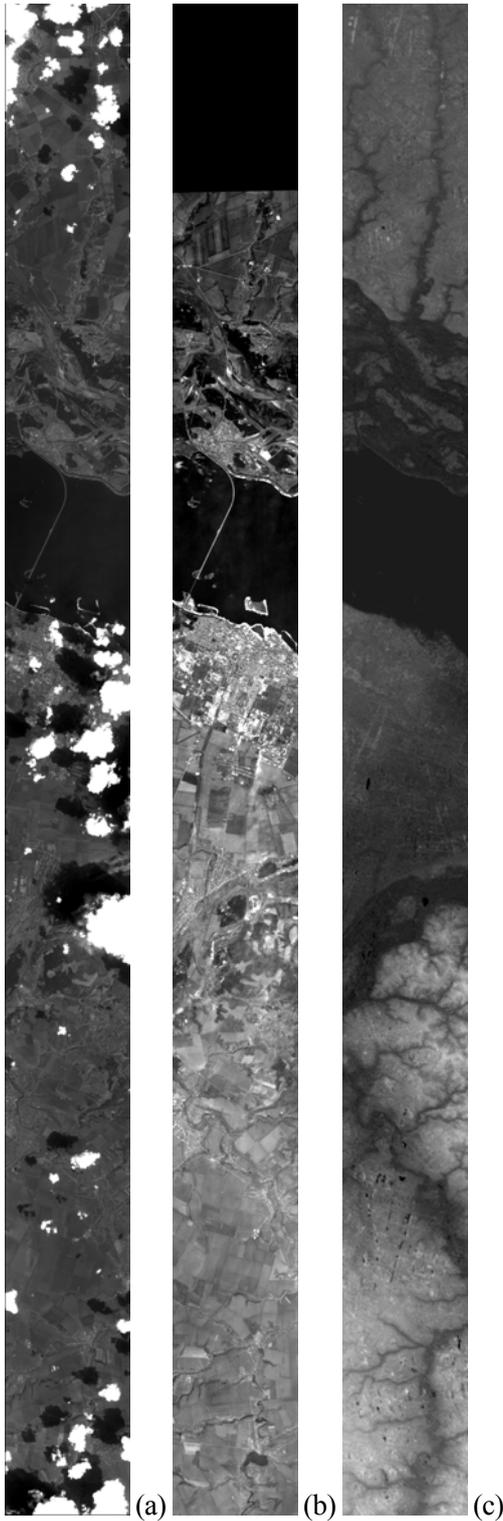

(a)    (b)    (c)

Fig. 4. Registered Hyperion band #25 (a), Landsat 8 band B1 (b) and ASTER GDEM (c). Gray levels ranging from black to white cover the intensity ranges 1100…3800 for Hyperion, 8500…9600 for Landsat 8 and 50…250m for ASTER GDEM. Images size is 256 by 3129 pixels.

Modifications of $CRLB_{fBm}$ bound allowing it to deal with such a complex noise model were introduced in our recent paper [27]. Using the methods proposed in [46-48], the following estimates were obtained (at the main scale of each image): $\sigma_c = 0$ pixels, $\sigma_n^2 = 69.64 + 0.071 \cdot I$ and $\sigma_n^2 = 69.6363 + 0.0714 \cdot I$ for Hyperion bands #25 and #155, respectively; $\sigma_c = 0.57$ pixels, $\sigma_n^2 = 35.55 + 0.021 \cdot I$, $\sigma_n^2 = 81.71 + 0.013 \cdot I$ and $\sigma_n^2 = 68.80 + 0.033 \cdot I$ for Landsat 8 bands B1, B5 and B8, respectively; and $\sigma_c = 1.2$ pixels, $\sigma_n^2 = 37.23 + 0.316 \cdot I^2$ and $\sigma_n^2 = 0.250 \cdot I^2$ for SIR-C radar images pr41419 and pr43020, respectively.

Noise in DEM images (data) is of different nature and mainly relates to elevation measurement error. Therefore, $\sigma_n$ is measured in meters in this case. We have assumed pure additive model for the DEM measurements error that follows Gaussian distribution and exhibits spatial correlation. The following estimates were obtained: $\sigma_c = 1.64$ pixels (at resolution of 30m), $\sigma_n^2 = 9.45\text{m}^2$. The obtained value $\sigma_n = 3.07\text{m}$ is consistent with mean SD of ASTER GDEM error of 3.52m reported by K. Becek in [49].

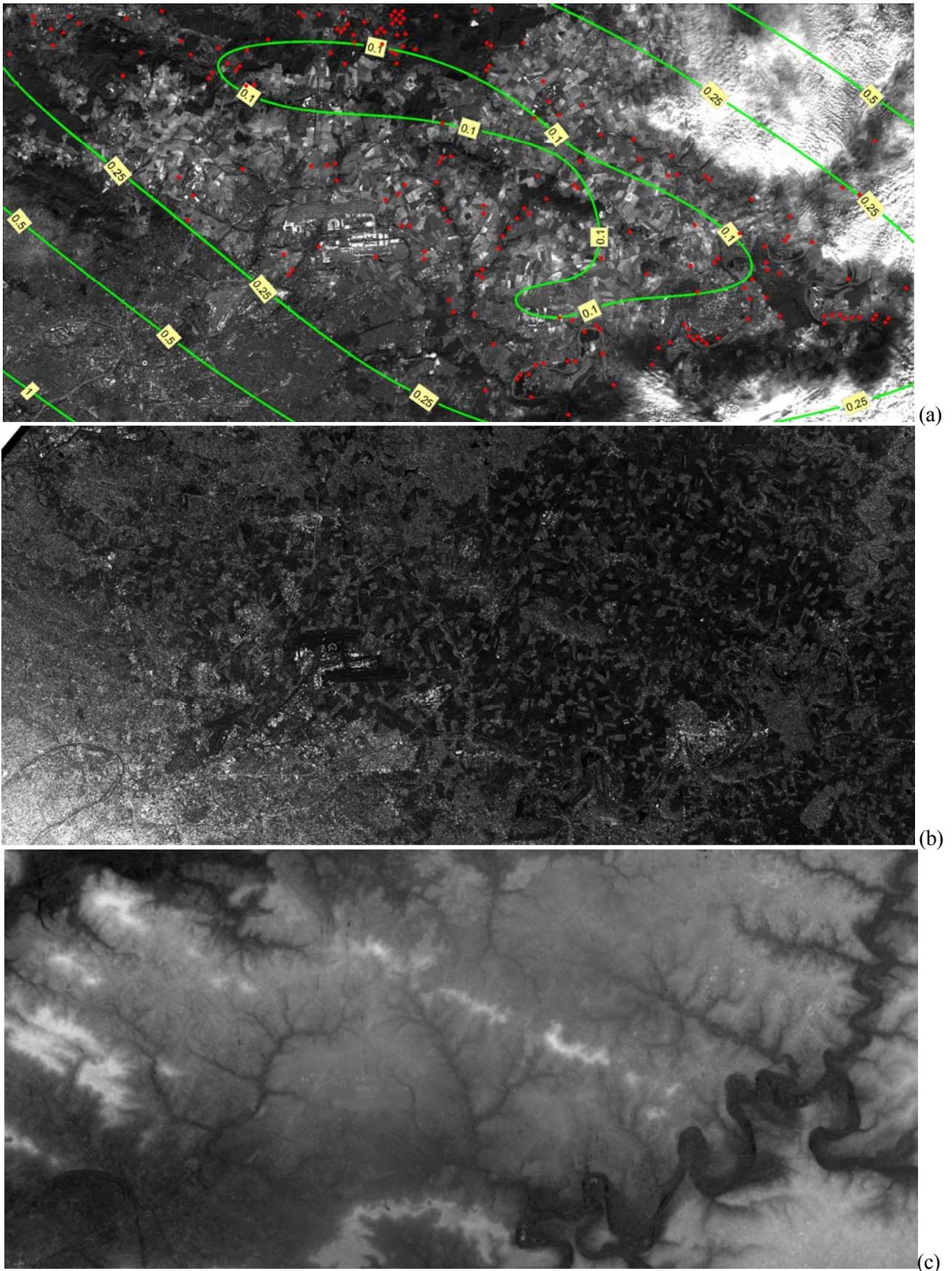

Fig. 5. Registered reference and template images for the test cases 3 and 4: Landsat 8 (a), SIR-C (b) and ASTER GDEM (c) images. Gray levels ranging from black to white cover the intensity ranges 5800… 8000 for Landsat 8, 0…255 for SIR-C and 0…255m for ASTER GDEM. Images size is 2151 by 951 pixels. The curves of constant registration accuracy (pixels) and registered CPs found by the RAE method for the test case 3 and the $2^{nd}$ order polynomial model are shown in (d) as green curves and red dots.

Relief influence is a factor that might have an impact on the analysis of registration results. Landsat 8 images in the test cases are orthorectified and free from relief influence as well as ASTER GDEM images. Hyperion and SIR-C images were corrected for relief influence based on sensor parameters, viewing angles and DEM information (using the same ASTER GDEM from the respective test cases).

**4.2. The RAE method performance analysis**

Let us first analyze performance of the RAE method and illustrate its behavior and distinctive features for the eight test cases. It is important to mention, that all results were obtained using the same RAE settings specified in the previous Section and $N_{TI} = N_{RI} = 17$ pixels.

The first feature – incremental registration accuracy improvement – is illustrated in Fig. 6. For the test cases 1…4 and affine registration model, the dependence of the mean registration error SD $\bar{\sigma}_{reg}$ is shown vs. the number of processed PCs, where $\bar{\sigma}_{reg}^2$ is defined as mean of $\sigma_{reg}^2(i_{TI}, j_{TI})$ over the whole reference image area. At the initialization stage, $\sigma_{reg}(i_{TI}, j_{TI})$ is set to $d_{\max 0}/3$ for all CFs. During initialization stage, $\bar{\sigma}_{reg}$ does not evolve until the first valid geometrical transform model parameters are found.

The complexity of the initialization stage depends on the registration problem complexity and it can be measured as the necessary number of PCs processed before completion. We can see that for the simplest case of optical to optical registration the initialization stage is very short, it takes only 75 PCs to complete (see Table 2 for numerical analysis results). Complexity increases for the multimodal test cases 2…4, which require from about 550 to 900 PCs to initialize geometrical transform parameters. Somewhat unexpectedly, the registrations of radar and optical images to DEM are simpler than optical to radar registration in terms of initialization stage complexity.

After the initial registration stage, $\bar{\sigma}_{reg}$ decreases fast due to having the search zone reduced and the increased probability of finding right correspondences. When registration error gets small enough (<1 pixel), the search zone becomes so narrow that the PC search procedure is no longer needed: majority of new PCs processed by the RAE method are true correspondences but their registration accuracy $\sigma_{PC.LB}$ can vary. In this operation mode, the mean registration error SD decreases approximately as reciprocal square root of the number of processed PCs. This stage of the RAE method can be viewed as a fine registration stage. The same observations were made from test cases 5…8 (see Table 3 for quantitative data) except that optical-to-DEM registration test case 6 has now the longest initialization stage.

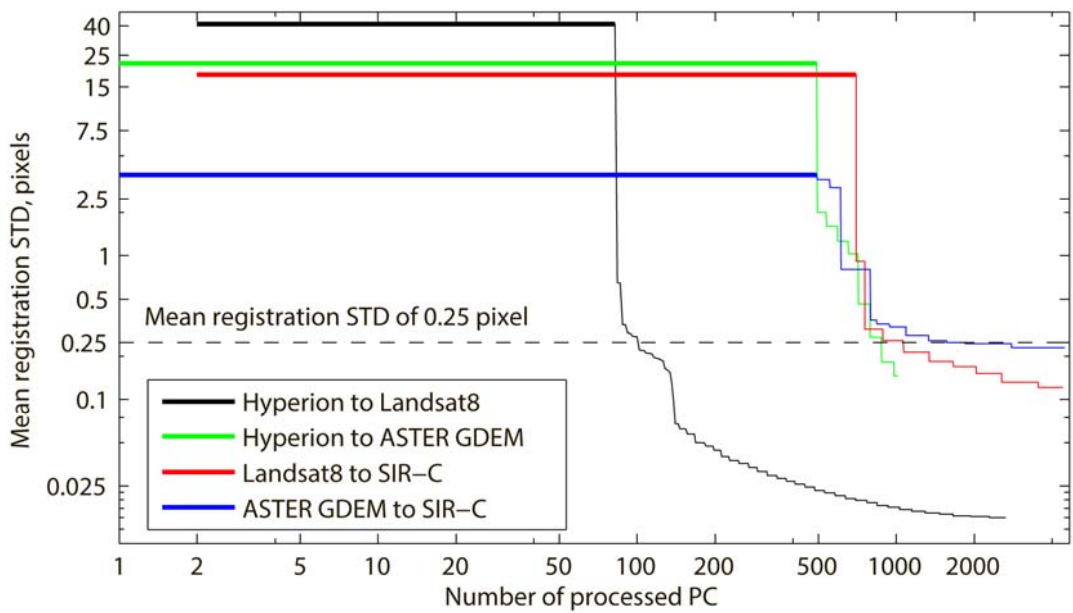

Fig. 6. RAE method registration accuracy improvement with iteration number for the four test cases

In practice, three distinctive stages of RAE operation can be outlined: the initialization stage where the true PCs search procedure is the most time consuming, the intermediate stage where the search zone is permanently decreased to reach the size of NCC lobe, and the fine registration stage where the search zone is virtually annihilated and the whole processing time is spent to increase the registration accuracy.

Qualitatively, the RAE method performance for the 1st and 2nd order polynomial model is summarized in Table 2 for TC1…TC4 and in Table 3 for TC5…TC8. The registration is successful in all cases. The number of registered CFs varies from 50 to about 1600 being larger for optical-to-optical and optical-to-radar cases and smaller for optical- and radar-to-DEM cases. This number is enough to provide subpixel registration over entire registered images area (as indicated by "Min/Mean/Max registration SD" line in Table 2) in all test cases with affine model. While this accuracy is normal for monomodal cases 1 and 4, ability to perform subpixel registration for the rest of multimodal cases outlines the strong capability of the RAE method to succeed in registering complex multimodal image pairs.

Table 2. REA method performance characteristics for the test cases 1…4

| Parameter | | Test case 1 | Test case 2 | Test case 3 | Test case 4 |
|---|---|---|---|---|---|
| Registration problem type | | optical-to-optical | optical-to-DEM | optical-to-radar | DEM-to-radar |
| Overall number of CF | | 2169 | 1224 | 9234 | 4321 |
| Percentage of inlying CF, $P_{CF}^{in}$ | | 95.1801 | 31.4109 | 64.7286 | 68.3713 |
| **Affine model** | | | | | |
| Length of the initialization stage | | 84 | 495 | 702 | 496 |
| Number of processed PC | | 2648 | 1019 | 4406 | 4473 |
| Percentage of inlying CF, $P_{CF}^{in}$ (init) | | 75.7917 | 5.8243 | 63.4428 | 21.5689 |
| Number of registered CF | | 1288 | 87 | 210 | 52 |
| Image area registered with SD less than 0.25 pixels, % | | 100 | 100 | 100 | 52.2183 |
| Min/Mean/Max registration SD | | 0.008/0.015/0.025 | 0.053/0.108/0.200 | 0.056/0.115/0.237 | 0.111/0.256/0.572 |
| RMSE (SD of absolute error), pixels (number of points) | $\sigma_{PC.LB} < 0.35$ | 0.5904 (1288) | 0.8436 (87) | 1.2529 (210) | 0.82621 (52) |
| | $\sigma_{PC.LB} < 0.225$ | 0.5080 (1186) | 0.7453 (64) | 0.8879 (24) | 0.46818 (6) |
| | $\sigma_{PC.LB} < 0.15$ | 0.3394 (956) | 0.4869 (26) | --- | --- |
| SD of normalized error, % | | 3.9043 | 4.2437 | 4.6807 | 3.0896 |
| **Second order polynomial model** | | | | | |
| Length of the initialization stage | | 157 | 2009 | 3140 | 1428 |
| Number of processed PC | | 2788 | 11991 | 7118 | 6453 |
| Percentage of inlying CF, $P_{CF}^{in}$ (init) | | 51.8646 | 1.9125 | 85.2335 | 63.0566 |
| Number of registered CF | | 1293 | 65 | 195 | 59 |
| Image area registered with SD less than 0.25 pixels, % | | 100 | 50.56 | 64.0626 | 30.1379 |
| Min/Mean/Max registration SD | | 0.013/0.024/0.062 | 0.086/0.298/0.998 | 0.092/0.267/1.395 | 0.158/0.491/2.364 |
| RMSE (SD of absolute error), pixels, x/y (number of points) | $\sigma_{PC.LB} < 0.35$ | 0.5921 (1293) | 0.62215 (65) | 1.065 (195) | 0.672 (59) |
| | $\sigma_{PC.LB} < 0.25$ | 0.5090 (1189) | 0.52815 (47) | 0.84431 (35) | 0.38554 (6) |
| | $\sigma_{PC.LB} < 0.15$ | 0.3371 (960) | 0.46619 (24) | --- | --- |
| SD of normalized error, $\sigma_{norm}$ | | 3.8549 | 3.3291 | 4.0768 | 2.4606 |

The most visible characteristic of registration complexity is percentage of inlying CFs, previously denoted $P_{CF}^{in}$. According to this criterion, the test cases 1…8 can be ordered in terms of increased complexity: optical-to-optical, optical-to-radar, DEM-to-radar, and the most complex case is optical-to-DEM registration. Recall that $P_{CF}^{in}$ defines the probability that there is, at least, one true correspondence for a CF that has, at least, one CRLB-validated PC. The value of $P_{CF}^{in}$ increases as more PCs are tested and it takes the lowest value at the end of initialization stage (these values for the 1st/2nd order models are 75.8/51.9% and 59.8/91.3% for optical-to-optical TCs 1 and 5; 5.8/1.9% and 0.5/2.4% for optical-to-DEM TCs 2 and 6; 63.4/85.2% and 47.9/31.8% for optical-to-radar TCs 3 and 7; 21.6/63.1% and 46.2/4.5% for radar-to-DEM TCs 4 and 8, see data in Tables 2 and 3). Among all test cases, the lowest value of $P_{CF}^{in}$ of about 2% (percentage of outliers about 98%) is obtained for the optical-to-DEM case. The RAE method succeeds even for this extremely low $P_{CF}^{in}$ value. To the best of our knowledge, there are no successful registration examples yet with percentage of outliers exceeding 90% for methods available in the literature.

Table 3. REA method performance characteristics for the test cases 5…8

| Parameter | | Test case 5 | Test case 6 | Test case 7 | Test case 8 |
|---|---|---|---|---|---|
| Registration problem type | | optical-to-optical | optical-to-DEM | radar-to-optical | radar-to-DEM |
| Overall number of CF | | 3307 | 10709 | 3396 | 2507 |
| Percentage of inlying CF, $P_{CF}^{in}$ | | 97.7874 | 39.0515 | 98.6447 | 82.0548 |
| **Affine model** | | | | | |
| Length of the initialization stage | | 22 | 5730 | 56 | 94 |
| Number of processed PC | | 2550 | 11911 | 5247 | 2561 |
| Percentage of inlying CF, $P_{CF}^{in}$ (init) | | 59.8048 | 0.5 | 47.9450 | 46.2770 |
| Number of registered CF | | 1337 | 434 | 1585 | 205 |
| Image area registered with SD less than 0.25 pixels, % | | 100 | 100 | 100 | 100 |
| Min/Mean/Max registration SD | | 0.010/0.019/0.034 | 0.028/0.064/0.130 | 0.013/0.024/0.045 | 0.046/0.097/0.194 |
| RMSE (SD of absolute error), pixels (number of points) | $\sigma_{PC.LB} < 0.35$ | 0.63388 (1337) | 0.85119 (434) | 0.72821 (1585) | 1.0139 (205) |
| | $\sigma_{PC.LB} < 0.225$ | 0.57903 (1256) | 0.81275 (328) | 0.5678 (1076) | 0.73816 (84) |
| | $\sigma_{PC.LB} < 0.15$ | 0.42427 (888) | 0.60457 (47) | 0.47738 (399) | 0.52898 (18) |
| SD of normalized error, % | | 4.1813 | 4.2451 | 3.6400 | 4.2591 |
| **Second order polynomial model** | | | | | |
| Length of the initialization stage | | 60 | 6162 | 66 | 2286 |
| Number of processed PC | | 2263 | 12840 | 5617 | 7157 |
| Percentage of inlying CF, $P_{CF}^{in}$ (init) | | 91.3188 | 2.4361 | 31.8714 | 4.5512 |
| Number of registered CF | | 1155 | 783 | 1594 | 192 |
| Image area registered with SD less than 0.25 pixels, % | | 100 | 100 | 100 | 94.4609 |
| Min/Mean/Max registration SD | | 0.016/0.036/0.113 | 0.033/0.052/0.162 | 0.020/0.035/0.103 | 0.069/0.140/0.392 |
| RMSE (SD of absolute error), pixels, x/y (number of points) | $\sigma_{PC.LB} < 0.35$ | 0.59135 (1155) | 0.75155 (783) | 0.74459 (1594) | 0.85119 (192) |
| | $\sigma_{PC.LB} < 0.225$ | 0.53047 (1089) | 0.72007 (651) | 0.55863 (1079) | 0.60198 (84) |
| | $\sigma_{PC.LB} < 0.15$ | 0.37541 (773) | 0.52004 (108) | 0.44259 (410) | 0.44945 (18) |
| SD of normalized error, $\sigma_{norm}$ | | 3.8297 | 3.8719 | 3.6166 | 3.5437 |

For affine model, RMSE of the found correspondences takes the smallest value of about 0.6 pixels for the optical-to-optical registration cases 1 and 5 and it is from 0.72 to 1.25 pixels for the test cases 2…4, 6…8. This RMSE is calculated using all found correspondences. As it was discussed above, each PC is characterized by its own registration accuracy and the RAE method is able to both characterize it and take it into account in its operation principle to improve registration performance. Let us next show the validity of $\sigma_{PC.LB}$ estimates for characterizing PCs registration accuracy at the local scale.

Fig. 7 shows 2D histogram of absolute registration errors (w.r.t. both horizontal and vertical directions) as a function of $\sigma_{PC.LB}$ for all correspondences found in the whole set of eight test cases. For a fixed $\sigma_{PC.LB}$ value, each histogram row is normalized to represent experimental p.d.f. of absolute registration error. The values of this error at the level $\pm 12\sigma_{PC.LB}$ are shown as white lines. These lines depict approximately the decision threshold separating inliers from outliers in the RAE method (recall that separation is achieved by comparing $\hat{P}_{kp}^{in}$ with the selected threshold 0.9; in turn, $\hat{P}_{kp}^{in}$ defined by (9) is dependent on $\sigma_{PC.LB}$). It is seen that the distribution of absolute errors concentrates more and more towards zero as $\sigma_{PC.LB}$ decreases. Moreover, for all $\sigma_{PC.LB}$ values, the registration error distribution remains non-uniform (close to normal) and decaying at the $\pm 12\sigma_{PC.LB}$ level. This behavior confirms that the observed concentration of registration errors is not due to inliers detection procedure but structural difference of registered image textures reflected by $\sigma_{PC.LB}$.

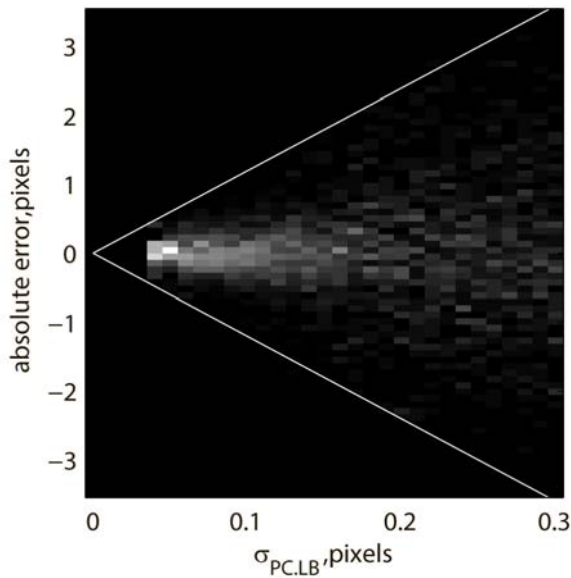

Fig. 7. 2D histogram of absolute registration error (w.r.t. both directions) and $\sigma_{PC.LB}$ for all correspondences found for the test cases 1-8. White color corresponds to higher probability density.

We can conclude, that $\sigma_{PC.LB}$ value reflects well the registration accuracy of the found correspondences and can be used to select the most accurate of them. This is demonstrated in Tables 2 and 3, where the registration results in terms of RMSE and upper threshold on $\sigma_{PC.LB}$ are given on the same line (at the bottom of Tables for each model). In all cases, the RMSE value of the found correspondences can be significantly reduced at the expense of getting fewer correspondences.

The normalized error values are obtained by dividing the registration error of each found correspondence by the respective value of $\sigma_{PC.LB}$. For an effective estimator, the normalized errors SD, denoted $\sigma_{norm}$, should be close to unity. For a real estimator, as the NCC considered in this study, the $\sigma_{norm}$ value and the estimator efficiency are related to each other as $e = 1/\sigma_{norm}^2$. For the NCC estimator, $\sigma_{norm}$ varies from 2.5 to 4.25 for all test cases (the lower value is obtained for more accurate 2$^{nd}$ order polynomial model as shown in Tables 2 and 3) which corresponds to efficiency value of about 10%. Therefore, we can justify again and validate the choice of the NCC estimator, selected in Section 3 just based on its efficiency observed in previous works [20, 26, 27]. We see in addition that the NCC estimator, though relatively simple, can be applied to a variety of complex multimodal registration problems with acceptable efficiency (such a possibility was mentioned by Mikolajczyk and Schmid in discussion Section in [30]).

The main differences between the registration results using the 2$^{nd}$ order polynomial model as compared to the 1$^{st}$ order model are as follows: RMSE of the registered PCs remains at the same level or slightly decreases due to more complex geometrical model. The initialization stage length tends to increase significantly (up to 20 times). The reason for this is that a more complex model has lower predictive capability, and tighter PC clustering is needed to initialize geometrical transform parameters with the RAE method. This also leads to reduced number of the registered PCs for the test cases 2 and 3 as some of the PCs do not belong to tight clusters. But if 2$^{nd}$ order model is significantly more adequate as compared to the 1$^{st}$ order model, the number of registered PCs can increase. This is the case for TC6. Overall registration process complexity increases as well due to the same reason: the search zone reduction is less effective for the 2$^{nd}$ order polynomial model with lower predictive capability. While the mean registration error SD remains almost the same, the maximal error increases significantly due to fast error divergence in regions not enclosed by the registered PCs.

Another feature of the RAE method that needs to be checked is its ability to predict registration accuracy at the global scale, that is correctness of the estimates $\sigma_{reg}(k)$. For all found correspondences, we have calculated the error between their position estimated by RAE and the one obtained using the reference transformation. We have normalized these errors by $\sigma_{reg}(k)$. For TCs 1…8, they lie approximately within ±6sigma interval with SD value about 1.5. SD of normalized errors exceeding unity means a slight underestimation of the registration error, but, overall, we can conclude that $\sigma_{reg}(k)$ is a correct estimate of the registration accuracy provided by the RAE method.

For test cases 1…4, five registered CFs corresponding to the lowest value of $\sigma_{PC.LB}$ are shown in Fig. 8. We see that for the optical-to-optical case, RI and TI CFs are almost identical. For the optical-to-DEM case, the intensities of RI and TI CFs are mostly inverted. Similarity between the pairs of CFs is lower than for test case 1 but is still obviously visible. For the optical-to-radar case, a very high level of speckle noise affecting radar image CFs is observed. For most of the pairs, the intensity inversion between the registered optical and radar images takes place as well. The registered fragments from DEM and radar images have very strong structural similarity. It is interesting to notice that many found correspondences do not belong to river basins, but to forest edges. The effect is that both radar and DEM images are sensitive to forest edges but in a different manner. The radar images reflect difference in surface roughness, the DEM images – in height.

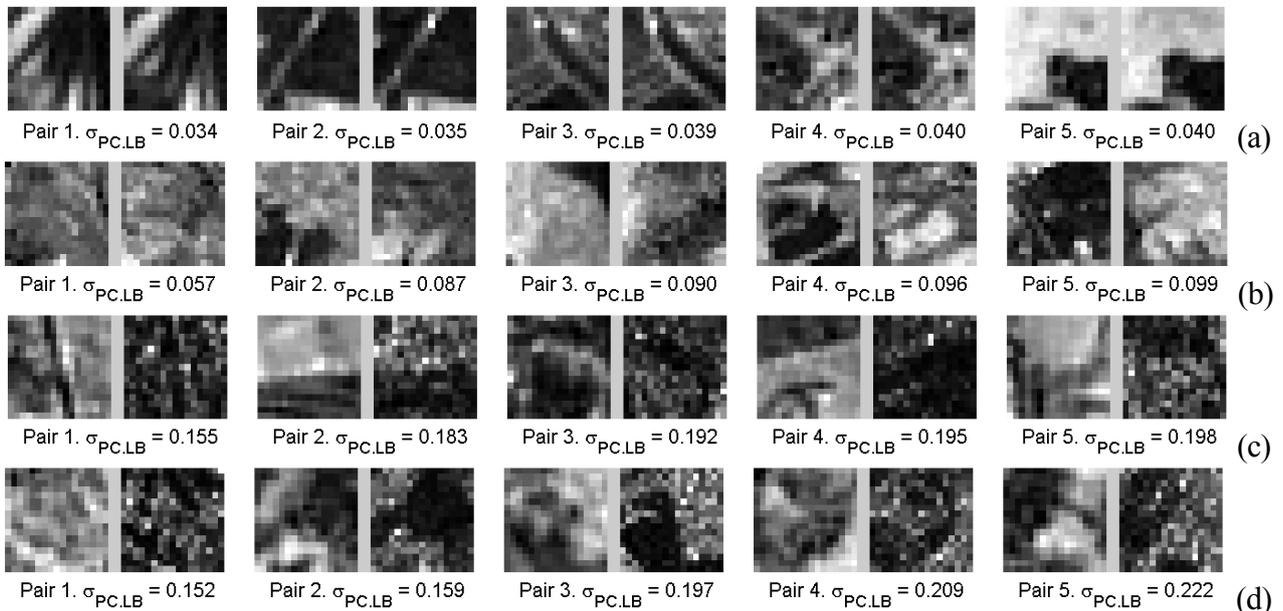

Fig. 8. Five control fragments with the lowest $\sigma_{PC.LB}$ value (given in pixels) found by the RAE method for test case 1-4: optical-to-optical (a), optical-to-DEM (b), optical-to-radar (c), and DEM-to-radar (d).

The pairs of the registered correspondences demonstrate complexity and variability of registration scenarios that can be successfully processed with the RAE method. An example of registration result by the RAE method for the test case 3 and the 2$^{nd}$ order polynomial model is shown in Fig. 5a (with presented curves of constant registration accuracy, $\sigma_{reg}$, and positions of the registered CFs).

**4.3. Comparative analysis**

The following registration methods were chosen for comparison: (1) the method for optical-to-radar images registration based on improved SIFT descriptor (referred to as ImprovedSIFT) [31], (2) the Locally Linear Transforming image registration method (referred to as LLT) [23], and (3) the registration based on MI similarity measure [50]. We have also tested a LLT method variant where SIFT descriptor is replaced with a more robust ImprovedSIFT descriptor (referred to as LLT+ImprovedSIFT). The comparison was done w.r.t. reference affine transformation. We stress that this reference transformation cannot be considered as very accurate since manual selection of control points for multimodal image pairs is subjective and inaccurate.

The code for LLT method is available from authors' webpage, the ImprovedSIFT method was implemented using VLfeat library, strictly following the original paper with one difference. We have found that among the three modifications of SIFT proposed in [31], namely (1) skipping the first octave, (2) skipping the step of preponderant orientation assignment, and (3) using multiple support regions, the first one does not improve registration quality for our test images. Therefore, we used only modifications 2 and 3 to obtain results shown below.

Registration based on MI similarity measure was implemented using Matlab's *imregister* function that includes MI calculated according to [50]. The *imregister* function allows finding local MI maxima using gradient descent algorithm [51] in the neighborhood of user-supplied starting guess. We have used the reference affine transformation as such a starting guess but with additional translation in the interval -50…50 pixels w.r.t. both vertical and horizontal translations (with step 2.5 pixels) such that one starting guess exactly corresponds to the reference affine transformation. For each starting guess, the registration has been performed w.r.t. affine transformation model and the result with maximal MI value has been taken as MI based registration output. Such a multistart approach also checks whether MI measure has global maxima close to the true registration parameters.

The performance of the ImprovedSIFT and LLT methods is comparatively evaluated based on the following criteria: the number of found correspondences, the number of true correspondences (with the registration error not exceeding 4 pixels), RMSE of the found correspondences, and mean and maximal registration error. The two latter criteria were calculated w.r.t. reference affine transformation. The MI-based registration accuracy is characterized only by the mean and maximal registration error. We did not use processing time as an additional criterion as our RAE method is one order of magnitude slower as compared to other analyzed methods.

The obtained quantitative results are presented in Tables 4 and 5 where successful registration outcomes are marked in bold. The LLT method with classical SIFT descriptor and LLT+ImprovedSIFT were able to register the simplest optical-to-optical TC1 and failed for TCs 2…7. ImprovedSIFT shows better results, it registered TC1 and multimodal radar-to-DEM TC8. But its accuracy and number of found correspondences are lower as compared to RAE method. MI demonstrates even better performance. For monomodal TC1, 5 and multimodal optical-to-radar TC3, 7, its performance is close to that of the RAE. For multimodal radar-to-DEM case TC 8, MI has global maxima close to the reference affine transformation but with higher mean/max errors as compared to RAE. For the most complex three TCs 2, 4 and 6, all involving DEM, the global MI maxima did not correspond to a solution close to the reference affine transformation and MI registration was unsuccessful.

Table 4. Performance of registration methods in comparison on test cases 1…4.

| | Method | Test case 1 optical to optical | Test case 2 optical to DEM | Test case 3 optical to radar | Test case 4 DEM to radar |
|---|---|---|---|---|---|
| RAE | Number of correspondences, all/true | **1288/1288** | **87/86** | **210/194** | **52/41** |
| | RMSE | **0.66** | **1.42** | **1.70** | **1.67** |
| | mean/max registration error | **0.64/1.41** | **1.557/2.913** | **1.9127/4.127** | **2.910/5.089** |
| Improved SIFT | Number of correspondences, all/true | **131/115** | 7/0 | 10/0 | 15/0 |
| | RMSE | **1.80** | 12.15 | 283.34 | 593.98 |
| | mean/max registration error | **0.73/1.74** | 24.10/43.23 | 403.83/481.50 | 825.71/851.99 |
| LLT | Number of correspondences, all/true | **60/53** | 13/0 | 10/0 | 11/0 |
| | RMSE | **1.99** | 430.15 | 868.03 | 578.67 |
| | mean/max registration error | **0.74/1.47** | 480.72/1182.66 | 912.27/1821.18 | 751.23/1551.9 |
| LLT+ Improved SIFT | Number of correspondences, all/true | **114/99** | 6/0 | 10/0 | 40/0 |
| | RMSE | **1.83** | 303.63 | 515.95 | 658.95 |
| | mean/max registration error | **0.82/1.91** | 338.29/981.46 | 820.53/1887.27 | 712.43/1769.7 |
| MI | mean/max registration error | **2.21/5.00** | 27.88/61.67 | **1.53/2.35** | 51.826/76.34 |

Table 5. Performance of registration methods in comparison on test cases 5…8.

| Method | | Test case 5 optical to optical | Test case 6 optical to DEM | Test case 7 optical to radar | Test case 8 DEM to radar |
|---|---|---|---|---|---|
| RAE | Number of correspondences, all/true | **1337/1337** | **407/344** | **1585/1575** | **205/197** |
| | RMSE | **0.76** | **2.13** | **1.26** | **1.71** |
| | mean/max registration error | **0.60/1.45** | **3.31/7.49** | **1.04/3.07** | **1.54/2.90** |
| Improved SIFT | Number of correspondences, all/true | **109/95** | 51/2 | 8/0 | **59/39** |
| | RMSE | **1.88** | 8.24 | 117.28 | **2.71** |
| | mean/max registration error | **1.21/2.79** | 15.11/27.28 | 152.55/195.59 | **3.62/7.60** |
| LLT | Number of correspondences, all/true | 23/0 | 11/0 | 15/0 | 9/0 |
| | RMSE | 1302.07 | 553.11 | 624.34 | 408.51 |
| | mean/max registration error | 1383.08/2960.88 | 858.49/1954.50 | 894.22/1850.68 | 517.06/1229.8 |
| LLT+ Improved SIFT | Number of correspondences, all/true | 13/5 | 25/0 | 71/0 | 4/2 |
| | RMSE | 829.61 | 786.87 | 566.98 | 375.23 |
| | mean/max registration error | 1010.67/2517.15 | 880.48/2131.65 | 448.62/940.35 | 483.73/951.22 |
| MI | mean/max registration error | **0.86/1.38** | 63.01/152.42 | **1.53/3.50** | 4.42/6.45 |

We conclude that the proposed RAE method is superior in terms of number of found correspondences and registration accuracy as compared to the set of state-of-the-art methods assessed here. It is able to cope with very complex registration scenarios where other methods in comparison fail to provide correct results.

## 5. CONCLUSION

In this paper, a new fully automatic area-based registration method has been proposed and proved suitable for a wide variety of remote sensing applications including such complex multimodal scenarios as registration of optical image to DEM, optical to radar images and DEM to radar image. The main features of the proposed method are its ability to quantify registration accuracy, deal with both linear and nonlinear geometrical models, reach linearithmic complexity with respect to image area (number of control fragments available), and attain compromise between processing time and area-based method accuracy.

The registration accuracy has been emphasized and used at both local and global scales for determining correspondences between registered images (control fragments) and for estimating geometrical transformation model parameters, respectively.

Unlike previous studies in the image registration field that mainly characterize *a posteriori* registration accuracy of correspondences at the output of the registration process, we have derived and next introduced the knowledge on local registration accuracy as an additional *a priori* information in the registration process. Such a possibility essentially comes from our previous efforts to quantify potential registration accuracy of textural noisy images achievable by area-based registration methods: $CRLB_{fBm}$ bound. Having such additional information, we have been able to improve the efficiency of both the outlier detection stage and the geometrical transformation parameter estimation stage. The most important benefit of our approach is that the registration accuracy at the global scale can be evaluated as the covariance matrix of estimates of polynomial geometrical model coefficients. We would like to outline that this registration accuracy does not need any ground truth to be determined and characterizes individual pairs of registered images taking into account their inherent structure. The validity of registration accuracy estimates at both scales has been experimentally confirmed. To the best of our knowledge, such a result was not published previously in the literature.

Local parametric image texture model and complex noise model (spatially correlated signal-dependent model) exploited within the proposed RAE registration method make it flexible and applicable to a wide range of registration problems including well studied optical-to-optical and optical-to-radar cases and scarcely studied DEM-to-optical and DEM-to-radar scenarios.

Comparison with state-of-the-art methods has shown that the RAE method can handle the most complex registration cases where other methods fail to provide accurate and reliable results.

Future work is intended to reduce high computational complexity of the RAE method. For this purpose, either the use of a multiscale approach or the search for simpler approximations of $CRLB_{fBm}$ bound can be considered. Another interesting direction could be in the use of more advanced similarity measures than NCC, e.g., mutual information measure or even feature-based descriptors.